\def\isarxiv{1} 

\ifdefined\isarxiv
\documentclass[11pt]{article}

\usepackage[numbers]{natbib}

\else
\documentclass{article}
\usepackage{neurips_2022}
\fi

\usepackage{amsmath}
\usepackage{amsthm}
\usepackage{amssymb}
\usepackage{algorithm}
\usepackage{subfig}
\usepackage{algpseudocode}
\usepackage{graphicx}
\usepackage{grffile}
\usepackage{wrapfig,epsfig}
\usepackage{url}
\usepackage{xcolor}
\usepackage{epstopdf}

\usepackage{bbm}
\usepackage{dsfont}

\allowdisplaybreaks

\ifdefined\isarxiv

\usepackage{tikz}
\usepackage{hyperref}  
\hypersetup{colorlinks=true,citecolor=blue,linkcolor=blue} 
\usetikzlibrary{arrows}
\usepackage[margin=1in]{geometry}

\else

\usepackage{microtype}
\usepackage{hyperref}
\definecolor{mydarkblue}{rgb}{0,0.08,0.45}
\hypersetup{colorlinks=true, citecolor=mydarkblue,linkcolor=mydarkblue}

\fi
\graphicspath{{./figs/}}

\newtheorem{theorem}{Theorem}[section]
\newtheorem{lemma}[theorem]{Lemma}
\newtheorem{definition}[theorem]{Definition}

\newtheorem{fact}[theorem]{Fact}

\newtheorem{claim}[theorem]{Claim}
\newtheorem{example}[theorem]{Example}

\newcommand{\wt}{\widetilde}

\newcommand{\R}{\mathbb{R}}

\newcommand{\A}{\mathsf{A}}

\DeclareMathOperator*{\E}{{\mathbb{E}}}

\DeclareMathOperator*{\Z}{\mathbb{Z}}

\DeclareMathOperator{\poly}{poly}

\DeclareMathOperator{\diag}{diag}

\DeclareMathOperator{\vect}{vec}

\makeatletter
\newcommand*{\RN}[1]{\expandafter\@slowromancap\romannumeral #1@}
\makeatother

\usepackage{lineno}

\begin{document}

\ifdefined\isarxiv

\date{}

\title{The Expressibility of Polynomial based Attention Scheme}
\author{
Zhao Song\thanks{\texttt{zsong@adobe.com}. Adobe Research.}
\and
Guangyi Xu\thanks{\texttt{guangyix@umich.edu}. University of Michigan.}
\and
Junze Yin\thanks{\texttt{junze@bu.edu}. Boston University.}
}

\else

\title{Intern Project} 
\maketitle 
\fi

\ifdefined\isarxiv
\begin{titlepage}
  \maketitle
  \begin{abstract}
Large language models (LLMs) have significantly improved various aspects of our daily lives. They have served as the foundation for virtual assistants, streamlining information retrieval and task automation seamlessly. These models have impacted numerous domains, from healthcare to education, enhancing productivity, decision-making processes, and accessibility. As a result, they have influenced and, to some extent, reshaped people's lifestyles. However, the quadratic complexity of attention in transformer architectures poses a challenge when scaling up these models for processing long textual contexts. This issue makes it impractical to train very large models on lengthy texts or use them efficiently during inference. While a recent study by \cite{kmz23} introduced a technique that replaces the softmax with a polynomial function and polynomial sketching to speed up attention mechanisms, the theoretical understandings of this new approach are not yet well understood.

In this paper, we offer a theoretical analysis of the expressive capabilities of polynomial attention. Our study reveals a disparity in the ability of high-degree and low-degree polynomial attention. Specifically, we construct two carefully designed datasets, namely $\mathcal{D}_0$ and $\mathcal{D}_1$, where $\mathcal{D}_1$ includes a feature with a significantly larger value compared to $\mathcal{D}_0$. We demonstrate that with a sufficiently high degree $\beta$, a single-layer polynomial attention network can distinguish between $\mathcal{D}_0$ and $\mathcal{D}_1$. However, with a low degree $\beta$, the network cannot effectively separate the two datasets. This analysis underscores the greater effectiveness of high-degree polynomials in amplifying large values and distinguishing between datasets. Our analysis offers insight into the representational capacity of polynomial attention and provides a rationale for incorporating higher-degree polynomials in attention mechanisms to capture intricate linguistic correlations.

  \end{abstract}
  \thispagestyle{empty}
\end{titlepage}

{\hypersetup{linkcolor=black}
\tableofcontents
}
\newpage

\else

\begin{abstract}

\end{abstract}

\fi

\section{Introduction}

The rapid progress of large language models (LLMs) like 
Llama \cite{tli+23,rgg+23}, 
GPT-1 \cite{rns+18}, 
GPT-3.5, 
Transformer \cite{vsp+17}, 
BERT \cite{dclt18}, 
PaLM \cite{cnd+22}, 
Llama 2 \cite{tms+23} 
GPT-2 \cite{rwc+19}, 
OPT \cite{zrg+22}, 
GPT-3 \cite{bmr+20}, 
Bard, 
GPT-4 \cite{o23}, 
has enabled powerful natural language capabilities. These models were trained on extensive text data, which enabled them to interpret the structures and patterns of human languages. The utilities of these LLMs are found across various domains, such as text generation, enhancing human-computer interaction, natural language understanding, multilingual communication, translation, and rapid prototyping. The key computational mechanism empowering their ability to understand language is the attention mechanism. Specifically, given an input sentence represented as vectors, attention finds pairwise correlations by computing inner products between the vector embeddings of words. Mathematically, we define the softmax attention mechanism as follows:
\begin{definition}[Self-Attention with Softmax Units]
    Let $g: \R \to \R$ be defined as $g(z) = \exp(z)$, where $g(W)_{i, j} = g(W_{i, j})$ if $W$ is a matrix. Given $A \in \R^{n \times d}$ and weights $Q,K, V \in \R^{d \times d}$, the attention computation can be written as 
\begin{align*}
    \mathsf{Att}(A,Q,K,V) = \underbrace{ D^{-1} }_{n \times n} \underbrace{ g( A QK^\top A^\top ) }_{n \times n} \underbrace{ A }_{n \times d} \underbrace{ V }_{d \times d}
\end{align*}
where $D:= \diag( g(AQ K^\top A^\top) {\bf 1}_n )$ where ${\bf 1}_n \in \R^n$ is a length-$n$ vector where all the entries are ones.
\end{definition}

However, the full self-attention results in a quadratic $O(n^2)$ complexity in a sequence, with length $n$. Under Strong Exponential time Hypothesis, there is no subquadratic time to compute the attention approximately \cite{as23}.   Therefore, the models are limited when they handle long contexts. There are various modified versions of the original transformer to mitigate this limitation, including the works of \cite{cld+20,kvpf20,wlk+20}. These efficient transformer variants aim to address the quadratic time complexity issue, allowing them to handle longer context lengths. Although many of these efficient transformer designs theoretically achieve linear training latency per step relative to context length, the practical improvements in the training latency have been underwhelming. Furthermore, \cite{lwl22} shows that some of these approaches have resulted in a reduction in model quality and most of the state-of-the-art models still rely on the original vanilla transformer.

Specifically, \cite{cld+20} introduces a kernel-based Transformer called Performer, which can provably approximate the softmax attention matrix. They propose and study the Fast Attention Via Positive Orthogonal Random features (FAVOR+) mechanism, where the query and key vectors, denoted as $Q_{i, *}$ and $K_{j, *}$ respectively, can be transformed through a random non-linear mapping into nonnegative vectors $\wt{Q}_{i, *}$  and $\wt{K}_{j, *}$, each of which has the dimension of $m = \Theta(d \log d)$, where $Q_{i, *}, K_{j, *} \in \R^d$ are the $i$-th row and $j$-th rows of the matrices $Q$ and $K$. This transformation holds with a high probability over the random mapping. The matrix $B$ and $\wt{B}$ with
\begin{align*}
    \wt{B}:= & ~ \wt{D}^{-1} \wt{C} , ~~~ \wt{D} := \diag( \wt{C} {\bf 1}_n ) , ~~~\wt{C} :=  W \circ( A \wt{Q} \wt{K}^\top A^\top ) ,  \\
    B := & ~ D^{-1} C, ~~~ D := \diag( C {\bf 1}_n ), ~~~ C:= W \circ \exp(A Q K^\top A^\top )
\end{align*}
satisfy that $\| B - \wt{B} \|_{\infty}$ is small and $W \in \{0, 1\}^{n \times n}$ is the mask matrix. Nevertheless, \cite{kmz23} believe there are two severe drawbacks: first, FAVOR+ is based on the assumption that $\ell_2$ norms of the query and key vectors are bounded; second, the result only holds for the fixed set of query and key vectors. In other words, a single randomized mapping may not preserve the attention matrix for all possible inputs, even when assuming bounded norms. 

Therefore, to solve this issue, \cite{kmz23} investigates the utilization of polynomials as an alternative to the softmax function in the formulation of attention mechanisms.

\begin{definition}[Self-Attention with Polynomial Units]\label{def:attention_poly_unit}
Let $g: \R \to \R$ be defined as $g(z) = z^{\beta}$ for $\beta \geq 2$, where $g(W)_{i, j} = g(W_{i, j})$ if $W$ is a matrix and $g(x)_{i} = g(x_{i})$ if $x$ is a vector. Given $A \in \R^{n \times d}$ and weights $Q,K, V \in \R^{d \times d}$, the attention computation can be written as 
\begin{align*}
    \mathsf{Att}(A,Q,K,V) = \underbrace{ D^{-1} }_{n \times n} \underbrace{ g( A QK^\top A^\top ) }_{n \times n} \underbrace{ A }_{n \times d} \underbrace{ V }_{d \times d}
\end{align*}
where $D:= \diag( g(AQ K^\top A^\top) {\bf 1}_n )$ where ${\bf 1}_n \in \R^n$ is a length-$n$ vector where all the entries are ones. 
\end{definition}

While \cite{kmz23} proposed using polynomials as an alternative activation in attention to enhance efficiency and model more complex correlations--$\mathsf{PolySketchFormer}$ replaces softmax units with higher-degree polynomial units in self-attention--the theoretical understanding of polynomial attention remains limited.

\subsection{Main Result}

In this paper, we provide a formal analysis of the expressive power of polynomial attention. We construct two carefully designed datasets and prove that high-degree polynomial attention can separate them but low-degree attention cannot. This establishes an expressivity gap between high-degree and low-degree polynomial attention.

\begin{theorem}[Main result, informal of Theorem~\ref{thm:main:formal}]\label{thm:main:informal}
There are two datasets $\mathcal{D}_0$ and $\mathcal{D}_1$, where $\mathcal{D}_1$ contains a feature with a larger value. 

For sufficiently large degree $\beta$, a single layer polynomial attention network can distinguish $\mathcal{D}_0$ and $\mathcal{D}_1$.

With low degree $\beta$, a single layer polynomial attention network cannot distinguish $\mathcal{D}_0$ and $\mathcal{D}_1$.
\end{theorem}

We illustrate the effectiveness of high-degree polynomials in amplifying large values, enabling the separation of datasets. Our theoretical analysis provides insights into the representational capacity of polynomial attention. The results justify the utilization of higher-degree polynomials in attention mechanisms to capture complex linguistic correlations.

\paragraph{Roadmap.}

In Section~\ref{sec:related_work}, we introduce the related work for this paper. In Section~\ref{sec:preli}, we present the basic notations and facts. In Section~\ref{sec:property_toy}, we analyze the properties of the datasets. In Section~\ref{sec:toy_result}, we give an overview of the theoretical performance of different models under the binary classification tasks. In Section~\ref{sec:app_self_attention}, we present the self-attention dataset and analyze its properties.


\section{Related Work}
\label{sec:related_work}

In this section, we give an overview of the related literature. 

\paragraph{Theoretical LLMs.}

In recent years, there has been a trend towards developing ever larger neural network models for natural language processing tasks. Many studies have analyzed the convergence properties and computational efficiency of exponential and hyperbolic activation functions in over-parameterized neural networks, as seen in \cite{lsx+23,ssz23,csy23b,csy23a,smk23}. Another line of research explores the capabilities and knowledge of LLMs, as evidenced by \cite{ll21,wwz+22,mbab22,byks22,ddh+21,ag23,hbkg23,zpga23,yjs+23,fpg+23,zwl+23,g23}. There are also studies dedicated to the multi-task training of LLMs, such as those found in \cite{lyb+22,xqp+22}, and fine-tuning \cite{mgn+23,psza23}. Additionally, optimization problems related to LLMs are analyzed in \cite{dgtt23,clmy21,wzz+23,rsm+23,ctl22,glha23}. Moreover, there is research that utilizes sketching techniques to speed up algorithms, like in \cite{syyz23_dp,qsw23,qsz23,qsy23,syyz23_linf,ssz23,rsz22,qjs+22,swyz23,qszz23}.

\paragraph{Theory and Application of Transformer.}

The Transformer architecture, consisting of an encoder-decoder structure with multiple layers of attention mechanisms as proposed by \cite{vsp+17}, has had a significant impact on natural language processing. Unlike previous works, such as \cite{hs97} which focused on recurrent neural networks and \cite{gag+17} which studied convolutional neural networks, the Transformer relies entirely on attention mechanisms to capture dependencies between input and output. The original Transformer model achieved state-of-the-art results in machine translation tasks. Subsequent research has extended the Transformer to address various language-related tasks. For instance, in the realm of pre-trained language representation, \cite{dclt18} investigated the BERT model, while \cite{rns+18} explored the GPT model, utilizing the Transformer decoder and encoder respectively during pre-training. For text generation, \cite{rwc+19} delved into the GPT-2 model, employing a causal Transformer decoder. Additionally, \cite{rsr+20} analyzed the T5 model, which frames all NLP tasks in a text-to-text format suitable for Transformer-based sequence-to-sequence modeling. Furthermore, the Transformer architecture has found applications in other modalities. The image transformer ViT \cite{dbk+20} divides an image into patches and applies self-attention directly. The speech transformer Speech-Transformer \cite{dxx18} operates on spectrogram slices, and Multimodal Transformers like ViLBERT \cite{lbpl19} utilize co-attention to combine image regions and text segments.

In addition to the extensive range of applications for Transformers, there is also substantial research dedicated to exploring the theoretical aspects of Transformers. One line of research focuses on optimization and convergence. 
\cite{zkv+20} underscored the significance of adaptive methods for attention models and introduced a novel adaptive approach for the attention mechanism.
\cite{szks21} explored how the single-head attention mechanism behaves in the context of Seq2Seq model learning. 
In the realm of neural networks, \cite{gms23} studied the convergence of over-parameterized neural networks employing exponential activation functions, addressing the challenge of over-parametrization.
On the algorithmic side, \cite{lsz23} presented an algorithm for regularized exponential regression, designed to operate efficiently with sparse input data. Their research demonstrated the algorithm's effectiveness across various datasets.

Another line of research focuses on studying in-context learners \cite{pmxa23}. \cite{gtlv22}, the focus was on training a model under in-context conditions, particularly for learning specific classes of functions like linear functions. The objective was to determine whether a model provided with information from specific functions within a class can effectively learn the ``majority'' of functions in that class through training. \cite{asa+22} demonstrated that these learners can implicitly execute traditional learning algorithms by continuously updating them with new examples and encoding smaller models within their activations. Additionally, \cite{onr+22} analyzed how Transformers function as in-context learners and by studying gradient descent, find connections between certain meta-learning formulations and the training process of Transformers in in-context tasks. 

More recently, research works also focus on interpreting the training procedure of multilayer Transformer architectures \cite{twz+23}, unlearning a subset of the training data \cite{er23}, task hinting \cite{ag23_hint}, hyper attention to reduce the computational challenge \cite{hjk+23}, and kernel-based method \cite{fhw+23}.

\paragraph{Attention.}

Attention mechanisms have been an important part of the Transformer architecture. The central idea behind transformer attention is ``self-attention'' where the input sequence attends to itself to compute representations of the sequence. Specifically, for each position in the input, the self-attention mechanism computes a weighted sum of all positions in the sequence, where the weights called "attention weights" are calculated by comparing the embedding at that position to all other positions using scaled dot-product attention. This allows the model to capture long-range dependencies in the sequences. Several improvements and variations of the self-attention mechanism have been proposed. For example, \cite{suv18} introduced relative position representations in the self-attention computation to better model position and distance relationships. \cite{sgbj19} proposed augmented memory transformers that add persistent memory slots to the encoder-decoder architecture to improve the memorization of previous information. Recent studies by \cite{zhdk23, clp+21, kkl20} have employed Locality Sensitive Hashing (LSH) techniques to approximate attention mechanisms. Particularly, \cite{zhdk23} introduced the ``KDEformer'', an efficient algorithm for approximating dot-product attention. This algorithm offers provable spectral norm bounds and outperforms various pre-trained models. \cite{clp+21} built $\mathsf{MONGOOSE}$, which is an end-to-end LSH framework used for neural network training. It adaptively performs LSH updates with a scheduling algorithm under the provable guarantees. \cite{kkl20} substituted dot-product attention with a method that employs locality-sensitive hashing for attention and substituted the conventional residual layers with reversible residual layers in order to improve the efficiency of Transformers. On the other hand, \cite{dls23,gsx23_incontext,gsy23_coin,gsyz23_quantum,dms23,syz23,zhdk23,as23,bsz23,swy23} had performed different kinds of the simplifications and variation of the attention computation. Finally, \cite{gswy23} used the tensor and Support Vector Machine (SVM) tricks to reformulate and solve the single-layer attention.

Attention has found wide-ranging applications across various domains. In the field of graph neural networks, \cite{vcc+17} explored these neural network architectures tailored for graph-structured data, calculating attention matrices between each node and its neighboring nodes. Within the context of the Transformer model \cite{vsp+17}, attention matrices were utilized to capture distinctions between words in a sentence. In the realm of image captioning, \cite{xbk+15} employed attention matrices to align specific image components with words in a caption.

\section{Preliminary}
\label{sec:preli}
Here in this section, we first introduce the mathematical notations being used in this paper. Then, in Section~\ref{sub:preli:prob}, we introduce the basic probability definitions and properties. In Section~\ref{sub:preli:def}, we present the important definition of the functions. In Section~\ref{sub:preli:binary}, we give the definition of the binary classification defined on datasets.

\paragraph{Notations.}

First, we introduce the notations for sets. We define $\Z_+ := \{1, 2, 3, \dots \}$. For a positive integer $n \in \Z_+$, the set $\{1,2,\cdots,n\}$ is denoted by $[n]$. We use $\R$ to denote the set of real numbers. For $n, d \in \Z_+$, we use $\R^d, \R^{n \times d}$ to denote the set of $d$-dimensional vectors and $n \times d$ matrices, whose entries are all in $\R$. For two sets $A_1$ and $A_2$, we use $A_1 \cup A_2$ to denote the union of $A_1$ and $A_2$, namely $A_1 \cup A_2 : = \{x \mid x \in A_1 \text{ or } x \in A_2\}$. For a sequence of sets $\{A_1, A_2, \dots, A_m\}$, we define $\cup_{i = 1}^m A_i := \{x \mid x \in A_i \text{ for some } i \in [m]\}$, where $m \in \Z_+ \cup \{\infty\}$. We use $A_1 \setminus A_2$ to denote the set $\{x \mid x \in A_1 \text{ and } x \notin A_2\}$.

Now, we introduce the notations related to vectors. Let $x, u \in \R^d$. For all $i \in [d]$, we define $x_i \in \R$ to be the $i$-th entry of $x$.
A vector of length $n$ with all entries being one is represented as ${\bf 1}_n$.
We use $\circ$ to denote the Hadamard product. We have $x \circ u \in \R^d$, where $(x \circ u)_{i} = x_i \cdot u_i$. If $u = x$, then we say $x \circ u = x^2$.

\begin{example}
    Let 
        $x = \begin{bmatrix}
        3\\
        4\\
        2
    \end{bmatrix}$.
    Let
        $y = \begin{bmatrix}
        5\\
        6\\
        7
    \end{bmatrix}$.

    Then, we have
    \begin{align*}
    x \circ y= 
    \begin{bmatrix}
        3\\
        4\\
        2
    \end{bmatrix} \circ
    \begin{bmatrix}
        5\\
        6\\
        7
    \end{bmatrix}
    =
    \begin{bmatrix}
        3 \cdot 5\\
        4 \cdot 6\\
        2 \cdot 7
    \end{bmatrix}
    =
    \begin{bmatrix}
        15\\
        24\\
        14
    \end{bmatrix}.
\end{align*}

Also, we have
\begin{align*}
    x^2 = \begin{bmatrix}
        3\\
        4\\
        2
    \end{bmatrix} \circ
    \begin{bmatrix}
        3\\
        4\\
        2
    \end{bmatrix} = 
    \begin{bmatrix}
        3 \cdot 3\\
        4 \cdot 4\\
        2 \cdot 2
    \end{bmatrix}
     = 
    \begin{bmatrix}
        9\\
        16\\
        4
    \end{bmatrix}.
\end{align*}
\end{example}

We use $\langle x, u \rangle$ to denote inner product, i.e., $\langle x, u \rangle := \sum_{i=1}^d x_i u_i$. We say a vector $x$ is $t$-sparse if $x$ has $t$ coordinates being nonzero, and other entries are all $0$.

Considering a matrix $A \in \R^{n \times d}$, its $i$-th column is referred to as $A_{*,i}$ for every $i \in [d]$. 
$A_{i, j} \in \R$ is an entry of $A$ at the $i$-th row and $j$-th column.
We use $\E[\cdot]$ to denote expectation. We use $\Pr[\cdot]$ to denote the probability. Let $f, g : \R \to \R$ be two functions. We use $g(n) = O(f(n))$ to denote that there exist positive constants $C$ and $x_0$ such that for all $n \geq x_0$, we have $|g(n)| \leq C \cdot f(n)$. We use $\poly(n)$ to denote a polynomial in $n$.

We use $A \in \R^{n \times d}$ to denote an $n \times d$ size matrix where each entry is a real number. We use $e_j$ to denote the unit vector where $j$-th entry is $1$ and others are $0$. For any positive integer $n$, we use $[n]$ to denote $\{1,2,\cdots, n\}$. For a matrix $A \in \R^{n \times d}$, we use $A_{i,j}$ to denote the an entry of $A$ which is in $i$-th row and $j$-th column of $A$, for each $i \in [n]$, $j \in [d]$. We use ${\bf 1}_{n \times n}$ to denote a $n \times n$ matrix where all the entries are ones. For a vector $x$ or a matrix $A$, we use $\exp(x)$ and $\exp(A)$ to denote the entry-wise exponential operation on them. For matrices $A \in \R^{n_1 \times d_1}$ and $B \in \R^{n_2 \times d_2}$, we use $A \otimes B \in \R^{n_1 n_2 \times d_1d_2}$ to denote a matrix such that its $((i_1 - 1)n_2 + i_2, (j_1 - i)d_2 + j_2)$-th entry is $A_{i_1,j_1} \cdot B_{i_2, j_2}$ for all $i_1 \in [n_1], j_1 \in [d_1], i_2 \in [n_2], j_2 \in [d_2]$. $\otimes$ is a binary operation called the Kronecker product. For a matrix $A \in \R^{n \times d}$, we use $\mathrm{vec}(A) \in \R^{nd}$ to denote the vectorization of $A$. Given two matrices $A,B \in \R^{n \times d}$, we also define the $\langle A, B \rangle := \langle \vect(A), \vect(B) \rangle$. We use $I_d$ to denote the $d \times d$ identity matrix. We use $A^\top$ to denote the transpose of a matrix $A$.

Using standard tensor-trick \cite{gsx23_incontext,gsy23_coin,gswy23,as23,as23_tensor}, we know that
\begin{fact}[Tensor-trick]\label{fac:tensor_trick}
Let $g: \R \rightarrow \R$ denote a function. 
Let $A_1, A_2 \in \R^{n \times d}$, let $Q \in \R^{d \times d}$, let $K \in \R^{d \times d}$, we have
\begin{align*}
\vect( g(A_1 Q K^\top A_2^\top) ) = g(\A \vect(QK^\top))
\end{align*}
where $\A:= A_1 \otimes A_2 \in \R^{n^2 \times d^2}$.
\end{fact}

Now, we introduce the linearity property of the inner product.

\begin{fact}\label{fac:vector}
    Let $a, b \in \R$.

    Let $x_1, y, x_2 \in \R^n$.

    Then, we have
    \begin{align*}
        a\langle x_1, y \rangle + b\langle x_2, y \rangle = \langle ax_1 + bx_2, y \rangle = \langle y, ax_1 + bx_2 \rangle = a\langle y, x_1 \rangle + b\langle y, x_2 \rangle.
    \end{align*}
\end{fact}

\subsection{Probability Tools}

In this section, we present some basic definitions and properties related to the probability theory. We first present the definition of Rademacher distribution.

\label{sub:preli:prob}

\begin{definition}
We say a random $x$ sampled from Rademacher distribution if $\Pr[x=+1]=1/2$ and $\Pr[x = -1] =1/2$.
\end{definition}

Now, we present two important probability properties: the union bound and the Hoeffding bound.

\begin{fact}[Union bound]\label{fac:union_bound}
    Given a collection of $n$ events $\{A_1, \cdots, A_n\}$.

    Then we have
    \begin{align*}
        \Pr[\cup_{i = 1}^n A_i] \geq 1 - \prod_{i = 1}^n (1 - \Pr[A_i]).
    \end{align*}
\end{fact}

\begin{lemma}[Hoeffding bound \cite{h63}]\label{lem:hoeffding_bound}
Let $X_1, \cdots, X_n$ denote $n$ independent bounded random variables in $[a_i,b_i]$. 

Let $X = \sum_{i=1}^n X_i$. 

Then we have
\begin{align*}
    \Pr[ | X - \E[X] | \geq t ] \leq 2 \exp \Big( - \frac{2t^2}{ \sum_{i=1}^n (b_i - a_i)^2 } \Big). 
\end{align*}
\end{lemma}

\subsection{Definitions of Functions}
\label{sub:preli:def}

In this section, we define the important functions.

\begin{definition}[degree-$\beta$ polynomial functions]\label{def:polynomial_functions}
Let $\beta > 0$ denote the degree.

We define $u_{\poly} : \R^{n \times d} \rightarrow \R^n$ as
\begin{align*}
    u_{\poly}(A; \beta ;x): = (A x)^{\beta},
\end{align*}
where the $i$-th entry of vector $(Ax)^{\beta} \in \R^n$ is $(Ax)_i^{\beta}$ for each $i \in [n]$.

We define $\alpha_{\poly}: \R^{n \times d} \rightarrow \R$ as
\begin{align*}
    \alpha_{\poly}(A; \beta; x) := \langle u_{\poly} (A; \beta; x) , {\bf 1}_n \rangle
\end{align*}

We define $f_{\poly} : \R^{n \times d} \rightarrow \R^n$ as
\begin{align*}
    f_{\poly}(A; \beta; x):= \alpha_{\poly}(A; \beta; x)^{-1} u_{\poly}(A;\beta;x)
\end{align*}
\end{definition}

\begin{definition}[ReLU]\label{def:relu}

We define $\phi: \R \to \R$ as
\begin{align*}
    \phi(z) := \max\{z,0\}.
\end{align*}

More generally, for parameter $\tau \in \R$, we define $\phi_{\tau} : \R \to \R$ as
\begin{align*}
    \phi_{\tau}(z): = \max\{ z - \tau, 0\}.
\end{align*}

\end{definition}

\begin{definition}\label{def:F}
Let $\tau > 0$ denote a parameter. 

Let $y \in \R^{n \times m}$. 

For each $j \in [m]$, we use $y_j \in \R^n$ to denote the $j$-th column of $y$. 

We define a three-layer neural network (with polynomial attention and ReLU activation) $F_{\poly}: \R^{n \times d} \rightarrow \R$ as
\begin{align*}
    F_{\poly}(A; \beta; x,y):= \phi( \sum_{j=1}^m \phi_{\tau}( \langle f_{\poly}(A; \beta; x) , y_j \rangle  )   )
\end{align*}

\end{definition}

\begin{definition}\label{def:y}
Let $C>1$ denote some constant. 

Let $m = C\log (n/\delta)$. 

Let $\delta \in (0,0.1)$ denote the failure probability.

Let $y \in \R^{n \times m}$. 

For each $j \in [m]$, we use $y_j$ to denote the $j$-th column of $y$. 

For each entry in $y_j$ we sampled it from Rademacher random variable distribution, i.e., where it's from $\{-1,+1\}$ with half probability.
\end{definition}

\subsection{Definition of Datasets for Binary Classification}
\label{sub:preli:binary}

In this section, we define the binary classification for the datasets. 

\begin{definition}[Binary classification]\label{def:binary_classification}
Let ${\cal D}_0$ and ${\cal D}_1$ be two sets of datasets.
\begin{itemize}
    \item For each $A \in \R^{n \times d}$ from ${\cal D}_0$, we assume that $(Ax)_i \in [2 ,4]$ for all $i \in [n]$.
    \item For each $A \in \R^{n \times d}$ from ${\cal D}_1$, we assume that there is one index $j \in [n]$ such that $(Ax)_j = 32 $ and for all $i \in [n] \backslash \{j\}$, we have $(Ax)_i \in [2,4]$.
\end{itemize}
\end{definition}

\section{Property of Dataset}
\label{sec:property_toy}

In this section, we analyze the mathematical properties of the dataset based on its definition presented earlier.

\begin{lemma}\label{lem:property_of_dataset_0}
Let $\sigma \sim \{-1,+1\}^n$ denote a random sign vector. 

Let $C > 1$ be a sufficiently large constant. 

Let $\delta \in (0,0.1)$. 

Let $\beta \in (0, 0.2 \log n)$.  

Then, for each $A \in {\cal D}_0$, we have
\begin{itemize}
    \item 
    Part 1. 
   \begin{align*}
        ~ \Pr_{\sigma 
        \sim \{-1,+1 \}^n }[ | \langle f_{\poly}(A; \beta; x) , \sigma \rangle | \leq 0.1 ] 
        \geq ~ 1- \delta/ \poly(n).
    \end{align*}
\end{itemize}
\end{lemma}

\begin{proof}

{\bf Proof of Part 1.}

We have that
\begin{align}\label{eq:alpha_poly_bound}
 \alpha_{\poly}(A; \beta; x) 
 = & ~ \langle u_{\poly}(A; \beta; x) , {\bf 1}_n \rangle \notag\\
 = & ~ \langle (A x)^{\beta} , {\bf 1}_n \rangle \notag\\
 \geq & ~ n \cdot 2^{\beta},
\end{align}
where the first step follows from the definition of $\alpha_{\poly}$ (see Definition~\ref{def:polynomial_functions}), the second step follows from the definition of $u_{\poly}(A; \beta; x)$ (see Definition~\ref{def:polynomial_functions}), and the third step follows from the fact that $A x \in \R^n$ and each entry $(A x)_i$ of the vector $A x$ is defined to be in the interval $[2, 4]$ (see Definition~\ref{def:binary_classification}).

By definition, we know that
\begin{align}\label{eq:upoly_bound}
 u_{\poly}(A; \beta; x)_i  
 = & ~  (A x)^{\beta}_i  \notag\\
 \leq & ~ 4^{\beta},
\end{align}
where the first step follows from the definition of $u_{\poly}$ (see Definition~\ref{def:polynomial_functions}), the second step follows from Definition~\ref{def:binary_classification}.

Note that we can show that
\begin{align*}
     f_{\poly}(A; \beta; x)_i  = & ~ \alpha_{\poly}(A; \beta; x)^{-1} \cdot  u_{\poly}(A; \beta; x)_i  \\
    \leq & ~ (n \cdot 2^{\beta})^{-1} \cdot ( 4^{\beta} ) \\
    = & ~ \frac{2^{\beta}}{n}
\end{align*}
where the first step follows from the definition of $f_{\poly}$ (see Definition~\ref{def:polynomial_functions}), the second step follows from Eq.~\eqref{eq:alpha_poly_bound} and Eq.~\eqref{eq:upoly_bound}, and the last step follows from simple algebra.

Therefore, we have
\begin{align*}
    f_{\poly}(A;\beta;x)_i \sigma_i \in [- \frac{2^{\beta}}{n} , + \frac{ 2^{\beta} }{n} ]
\end{align*}
We also know that
\begin{align}\label{eq:expectation_f_poly}
    \E[ f_{\poly}(A;\beta;x)_i \sigma_i ] = 0, ~~~ \forall i \in [n]
\end{align}

For simplicity, we let 
\begin{align}\label{eq:X}
    X := \sum_{i=1}^n f_{\poly}(A; \beta; x)_i \sigma_i  = \sum_{i=1}^n X_i.
\end{align}

By Eq.~\eqref{eq:expectation_f_poly}, we have
\begin{align}\label{eq:EX}
    \E[X] 
    = & ~ \E[\sum_{i=1}^n f_{\poly}(A; \beta; x)_i \sigma_i] \notag\\ 
    = & ~ \sum_{i=1}^n \E[ f_{\poly}(A; \beta; x)_i \sigma_i] \notag\\ 
    = & ~ \sum_{i=1}^n 0 \notag\\ 
    = & ~ 0,
\end{align}
where the first step follows from the definition of $X$ (see Eq.~\eqref{eq:X}), the second step follows from the linearity of expectation, the third step follows from Eq.~\eqref{eq:expectation_f_poly}, and the last step follows from simple algebra.

Therefore, by Hoeffding inequality (see Lemma~\ref{lem:hoeffding_bound}), we can get
\begin{align*}
    \Pr[|X - \E[X]| \geq  C \cdot \sqrt{ n \log(n/\delta) } \cdot \frac{ 2^{\beta} } {n}] \leq \delta /\poly(n),
\end{align*}
as we let $C$ to be sufficiently large.

Therefore, we have with at least $1-\delta /\poly(n)$ probability, 
\begin{align}\label{eq:X-EX}
    |X - \E[X]| \leq  C \cdot \sqrt{ n \log(n/\delta)} \cdot \frac{ 2^{\beta} } {n}
\end{align}

Therefore, with at least $1-\delta /\poly(n)$, we have
\begin{align*}
|\langle f_{\poly}(A;\beta;x) , \sigma \rangle |
  = & ~ | \sum_{i=1}^n f_{\poly}(A; \beta; x)_i \sigma_i | \\
  = & ~ |X|\\
  = & ~ |X - \E[X]|\\
 \leq & ~ C \cdot \sqrt{ n \log(n/\delta) } \cdot \frac{ 2^{\beta} } {n}   \\
 = & ~ C \cdot \frac{ \sqrt{ \log(n/\delta) } }{ \sqrt{n} }  \cdot 2^{\beta} \\
 \leq & ~ 0.1,
\end{align*}
where the first step follows from the definition of ${\bf 1}_n$, the second step follows from Eq.~\eqref{eq:X}, the third step follows from Eq.~\eqref{eq:EX}, the fourth step follows from Eq.~\eqref{eq:X-EX}, the fifth step follows from simple algebra, and the last step follows from the definition of $\beta$ (see from the Lemma statement).

\end{proof}

\begin{lemma}\label{lem:property_of_dataset_1}
Let $\sigma \sim \{-1,+1\}^n$ denote a random sign vector. 

Let $C > 1$ be a sufficiently large constant. 

Let $\delta \in (0,0.1)$. 

Then, for each $A \in {\cal D}_1$, we have
\begin{itemize}
    \item Part 1. if $\beta \in (0,0.01 \log n)$,  
    \begin{align*}
        \Pr_{\sigma \sim \{-1,+1\}^n }[ | \langle f_{\poly}(A;\beta;x) , \sigma \rangle | \leq C \frac{ \sqrt{ \log(n/\delta)} } {\sqrt{ n } } \cdot 16^{\beta} ] \geq 1- \delta/ \poly(n).
    \end{align*}
    \item Part 2. if $\beta \in (\frac{1}{3} \log n,  0.49 \log n)$ \begin{align*}
        \Pr_{\sigma \sim \{-1,+1\}^n }[ \langle f_{\poly}(A; \beta; x) , \sigma \rangle  \geq \frac{1}{3} ] \geq \frac{1}{4}.
    \end{align*}  
\end{itemize}
\end{lemma}
\begin{proof}

{\bf Proof of Part 1.}

We know that 
\begin{align}\label{eq:alpha_poly_bound_2}
\alpha_{\poly}(A; \beta; x) 
 = & ~ \langle u_{\poly}(A; \beta; x) , {\bf 1}_n \rangle \notag\\
 = & ~ \langle (A x)^{\beta} , {\bf 1}_n \rangle \notag\\
\geq & ~ (n-1) \cdot 2^{\beta} + 32^{\beta} \notag\\
\geq & ~ n \cdot 2^{\beta},
\end{align}
where the first step follows from the definition of $\alpha_{\poly}$ (see Definition~\ref{def:polynomial_functions}), the second step follows from the definition of $u_{\poly}$ (see Definition~\ref{def:polynomial_functions}), the third step follows from the fact that $A x \in \R^n$ and one entry of $A x$ is $32$ and other entries of $A x$ are defined to be in the interval $[2, 4]$ (see Definition~\ref{def:binary_classification}), and the last step follows from simple algebra.

Therefore, we can show that
\begin{align*}
     f_{\poly}(A; \beta; x)_i  = & ~ \alpha_{\poly}(A; \beta; x)^{-1} \cdot u_{\poly}(A; \beta; x)_i  \\ 
    \leq & ~  ( n 2^{\beta}  )^{-1} \cdot ( 32^{\beta} ) \\
    = & ~ \frac{ 16^{\beta} }{ n }  
\end{align*}
where the first step follows from the definition of $f_{\poly}$ (see Definition~\ref{def:polynomial_functions}), the second step follows from Definition \ref{def:binary_classification} and Eq.~\eqref{eq:alpha_poly_bound_2}, the third step follows from simple algebra.

Therefore, we have
\begin{align*}
    f_{\poly}(A;\beta;x)_i \sigma_i \in [- \frac{ 16^{\beta} }{ n }   , + \frac{ 16^{\beta} }{ n }   ]
\end{align*}
We also know that
\begin{align}\label{eq:expectation_fpoly}
   \E[ f_{\poly}(A;\beta;x)_i \sigma_i ] = 0, ~~~ \forall i \in [n]
\end{align}

For simplicity, we let
\begin{align}\label{eq:X_1}
    X := \sum_{i = 1}^n f_{\poly}(A;\beta;x)_i\sigma_i=\sum_{i = 1}^n X_i.
\end{align}

By Eq.~\eqref{eq:expectation_fpoly}, we have
\begin{align}\label{eq:EX_1}
    \E[X] = & ~ \E[\sum_{i = 1}^n f_{\poly}(A;\beta;x)_i \sigma_i] \notag\\
    = & ~ \sum_{i = 1}^n \E[f_{\poly}(A;\beta;x)_i \sigma_i] \notag\\
    = & ~ \sum_{i = 1}^n 0 \notag\\
    = & ~ 0,
\end{align}
where the first step follows from the definition of $X$ (see Eq.~\eqref{eq:X_1}), the second step follows from the linearity of expectation, the third step follows from Eq.~\eqref{eq:expectation_fpoly}, and the last step follows from simple algebra. 

Therefore, by Hoeddfing inequality (see Lemma~\ref{lem:hoeffding_bound}), we can get
\begin{align*}
    \Pr[|X - \E[X]| \geq C \cdot \sqrt{n\log(n/\delta)} \cdot \frac{16^\beta}{n}] \leq \delta/\poly(n),
\end{align*}
as we let $C$ to be sufficiently large.

Therefore, we have with at least $1 - \delta/\poly(n)$ probability,
\begin{align}\label{eq:X_EX_1}
     |X - \E[X]| \leq  C \cdot \sqrt{ n \log(n/\delta)} \cdot \frac{ 16^{\beta} } {n}
\end{align}

Therefore, with at least $1 - \delta/\poly(n)$, we have
\begin{align*}
    |\langle f_{\poly}(A;\beta;x) , \sigma \rangle |
  = & ~ | \sum_{i=1}^n f_{\poly}(A; \beta; x)_i \sigma_i | \\
  = & ~ |X|\\
  = & ~ |X - \E[X]|\\
 \leq & ~ C \cdot \sqrt{n \log(n/\delta)} \cdot \frac{16^{\beta}} {n}   \\
 = & ~ C \cdot \frac{\sqrt{\log(n/\delta)}}{\sqrt{n}}  \cdot 16^{\beta},
\end{align*}
where the first step follows from the definition of ${\bf 1}_n$, the second step follows from Eq.~\eqref{eq:X_1}, the third step follows from Eq.~\eqref{eq:EX_1}, the fourth step follows from Eq.~\eqref{eq:X_EX_1}, the fifth step follows from simple algebra, and the last step follows from the definition of $\beta$ (see from the Lemma statement).

{\bf Proof of Part 2.}

We know that 
\begin{align}\label{eq:alpha_poly_bound_3}
     \alpha_{\poly}(A; \beta; x) 
     = & ~ \langle u_{\poly}(A; \beta; x) , {\bf 1}_n \rangle \notag\\
    = & ~ \langle (A x)^{\beta} , {\bf 1}_n \rangle \notag\\
     \leq & ~ (n-1) \cdot 4^{\beta} + 32^{\beta} \notag\\
     \leq & ~ 8^{\beta} \cdot 4^{\beta} + 32^{\beta},
\end{align}
where the first step follows from the definition of $\alpha_{\poly}$ (see Definition~\ref{def:polynomial_functions}), the second step follows from the definition of $u_{\poly}$ (see Definition~\ref{def:polynomial_functions}), the third step follows from Definition~\ref{def:binary_classification}, and the last step follows from $8^{\beta} > n$ ($\beta > \frac{1}{3} \log n$).

There is one index $j \in [n]$, we know that
\begin{align*}
     f_{\poly}(A; \beta; x)_j  = & ~ \alpha_{\poly}(A; \beta; x)^{-1} \cdot  u_{\poly}(A; \beta; x)_j  \\
    \geq & ~ ( 32^{\beta} + 32^{\beta} )^{-1} \cdot ( 32^{\beta} ) \\
    \geq & ~ \frac{1}{2},
\end{align*}
where the first step follows from the definition of $f_{\poly}$ (see Definition~\ref{def:polynomial_functions}), the second step follows from Eq.~\eqref{eq:alpha_poly_bound_3} and Definition~\ref{def:binary_classification}, and the third step follows from simple algebra.

We also know that 
\begin{align}\label{eq:alphaa_poly_2}
     \alpha_{\poly}(A; \beta; x) 
     = & ~ \langle u_{\poly}(A; \beta; x) , {\bf 1}_n \rangle \notag\\
    = & ~ \langle (A x)^{\beta} , {\bf 1}_n \rangle \notag\\
     \geq & ~ (n-1) \cdot 2^{\beta} + 32^{\beta} \notag\\
     \geq & ~ n \cdot 2^{\beta},
\end{align}
where the first step follows from the definition of $\alpha_{\poly}$ (see Definition~\ref{def:polynomial_functions}), the second step follows from the definition of $u_{\poly}$ (see Definition~\ref{def:polynomial_functions}), the third step follows from Definition~\ref{def:binary_classification}, and the last step follows from simple algebra.

For all the $i \in [n] \backslash \{j\}$, we know that 
\begin{align*}
     f_{\poly}(A; \beta; x)_i  = & ~ \alpha_{\poly}(A; \beta; x)^{-1} \cdot  u_{\poly}(A,x)_i  \\
    \leq & ~ (n \cdot 2^{\beta} )^{-1} \cdot ( 4^{\beta} ) \\
    \leq & ~ \frac{2^{\beta}}{n},
\end{align*}
where the first step follows from the definition of $f_{\poly}$ (see Definition~\ref{def:polynomial_functions}), the second step follows from Definition~\ref{def:binary_classification} and Eq.~\eqref{eq:alphaa_poly_2}, and the last step follows from simple algebra.

For convenient, let $S = \{j \}$. We use $e_S \in \R^n$ to denote the vector the coordinates in $S$ are ones and coordinates not in $S$ are zeros.

Therefore, we have that (let us call this event E1)
\begin{align*}
 \langle f_{\poly}(A;\beta;x) , e_S \circ \sigma \rangle 
 = & ~ f_{\poly}(A;\beta;x)_j \sigma_j \\
 \geq & ~ \frac{1}{2}.
\end{align*}
happens with probability $1/2$.

The reason of the above equation is 
\begin{align*}
    f_{\poly}(A;\beta;x)_j \geq 1/2
\end{align*}
and 
\begin{align*}
    \Pr[\sigma_j = +1] = 1/2
\end{align*}
and
\begin{align*}
    \Pr[\sigma_j = -1] = 1/2.
\end{align*}

Using Hoeffding inequality, we have (let us call this event E2)
\begin{align*}
|\langle f_{\poly}(A;\beta;x) , e_{[n] \backslash S} \circ \sigma \rangle | 
\leq & ~ C \cdot \frac{ \sqrt{\log(n/\delta)} }{ \sqrt{n} } \cdot 2^{\beta} \\
\leq & ~ 0.1,
\end{align*}

with probability at least $1-\delta/\poly(n) \geq 1- 0.1$. Here, the last step follows from 
\begin{align*}
    \beta < 0.49 \log n.
\end{align*}

Therefore, we have that
\begin{align*}
  \langle f_{\poly}(A; \beta; x) ,  \sigma \rangle  
  \geq & ~ \frac{1}{2} - 0.1 \\
  \geq & ~ \frac{1}{3},
\end{align*}
where the second step follows from simple algebra.

Union bound the above two events (E1 and E2), the succeed probability 
is at least 
\begin{align*}
\frac{1}{2} - 0.1 \geq \frac{1}{4}.
\end{align*}
\end{proof}

\section{Binary Classification}\label{sec:toy_result}
In this section, we provide an overview of the theoretical analysis of the performance of different models in binary classification tasks. More specifically, in Section~\ref{sub:toy_result:high}, we analyze the property of the high degree polynomial attention; in Section~\ref{sub:toy_result:low}, we analyze the property of the low degree polynomial attention

\subsection{High-Degree Polynomial Attention}
\label{sub:toy_result:high}

Here, in this section, we start to analyze high-degree polynomial attention.

\begin{lemma}
If $\beta \in ( \frac{1}{3} \log n, 0.45 \log n)$ and $\tau = 0.2$, then

\begin{itemize}
\item For each data $A$ from ${\cal D}_0$, $F_{\poly}(A; \beta; x; y) = 0$ with probability at least $1-\delta/\poly(n)$.

\item For each data $A$ from ${\cal D}_1$, $F_{\poly}(A; \beta; x; y) > 0 $ with probability at least $1-\delta/\poly(n)$.
\end{itemize}
\end{lemma}
\begin{proof}

{\bf Proof of Part 1.}

Note that all the $y_{*,l}$ are independent, for each $l\in [m]$, we call Part 1 of Lemma~\ref{lem:property_of_dataset_0}, we can show that
\begin{align}\label{eq:phi_tau_0}
  \phi_{\tau}( \langle f_{\poly}(A; \beta; x) , y_{*,l} \rangle   ) 
  = & ~ \max\{  \langle f_{\poly}(A; \beta; x) , y_{*,l} \rangle - \tau, 0\} \notag\\
  = & ~ 0,
\end{align}
with probability $1-\delta/\poly(n)$, where the first step follows from the definition of $\phi_{\tau}$ (see Definition~\ref{def:relu}) and the second step follows from $| \langle f_{\poly}(A; \beta; x) , y_{*,l} \rangle | \leq 0.1$ (see from Part 1 of Lemma~\ref{lem:property_of_dataset_0}) and $\tau = 0.2$ (see from the Lemma statement).

We take a union bound (see Fact~\ref{fac:union_bound}) over all $l \in [m]$. 

Thus, we have with probability $1-m \cdot \delta/\poly(n) \geq 1-\delta/\poly(n)$,
\begin{align*}
F_{\poly}(A; \beta; x; y) 
= & ~ \phi( \sum_{l=1}^m   \phi_{\tau}( \langle f_{\poly}(A; \beta; x) , y_{*,l} \rangle   ) ) \\
= & ~ 0,
\end{align*}
where the first step follows from the definition of $F_{\poly}$ (see Definition~\ref{def:F}) and the second step follows from Eq.~\eqref{eq:phi_tau_0}.

{\bf Proof of Part 2.}

By Part 2 of Lemma~\ref{lem:property_of_dataset_1}, we have 
\begin{align*}
  \phi_{\tau}( \langle f_{\poly}(A; \beta; x) , y_{*,l} \rangle   ) 
  = & ~ \max \{\langle f_{\poly}(A; \beta; x) , y_{*,l} \rangle - \tau, 0\} \\
  > & ~ \max\{ 1/3 - \tau, 0\} \\
  = & ~ \max\{1/3-0.2, 0 \} \\
  \geq & ~ 1/10
\end{align*}
with probability $1/4$, where the first step follows from the definition of $\phi_{\tau}$ (see Definition~\ref{def:relu}), the second step follows from Part 2 of Lemma~\ref{lem:property_of_dataset_1}, the third step follows from $\tau = 0.2$ (see the Lemma statement), and the last step follows from simple algebra.

Since all $l \in [m]$ are independent, thus there exists one $l \in [m]$ such that 
\begin{align*}
  \phi_{\tau}( \langle f_{\poly}(A; \beta; x) , y_{*,l} \rangle   ) > 1/10
\end{align*}
the probability is $1-(3/4)^m \geq 1-\delta/\poly(n)$.

Thus, with probability $1-\delta/\poly(n)$, we have
\begin{align*}
F_{\poly}(A; \beta; x; y) = \phi( \sum_{l=1}^m   \phi_{\tau}( \langle f_{\poly}(A; \beta; x) , y_{*,l} \rangle   ) ) > 0
\end{align*}
holds. 
\end{proof}

\subsection{Low-Degree Polynomial Attention}
\label{sub:toy_result:low}

Now, in this section, we start to analyze low-degree polynomial attention.

\begin{lemma}
If $\beta \in (0, 0.01 \log n)$ and let $\tau = 0.2$. 
\begin{itemize}
\item {\bf Part 1.}
For each data $A$ from ${\cal D}_0$, $F_{\poly}(A;\beta; x; y) = 0 $ with probability at least $1-\delta/\poly(n)$.
\item {\bf Part 2.} For each data $A$ from ${\cal D}_1$, $F_{\poly}(A;\beta; x; y) = 0 $ with probability at least $1-\delta/\poly(n)$.
\end{itemize}
\end{lemma}
\begin{proof}
{\bf Proof of Part 1.}

Note that all the $y_{*,l}$ are independent, for each $l\in [m]$, we call Part 1 of Lemma~\ref{lem:property_of_dataset_0}, we can show that, with probability $1 - \delta / \poly(n)$,
\begin{align}\label{eq:phi_tau_0_2}
  \phi_{\tau}( \langle f_{\poly}(A; \beta; x) , y_{*,l} \rangle   ) 
  = & ~ \max \{\langle f_{\poly}(A; \beta; x) , y_{*,l} \rangle - \tau, 0 \} \notag\\
  = & ~ \max \{0.1 - 0.2, 0 \} \notag\\
  = & ~ 0,
\end{align}
where the first step follows from the definition of $\phi_{\tau}$ (see Definition~\ref{def:relu}), the second step follows from Part 1 of Lemma~\ref{lem:property_of_dataset_0} and the assumption that $\tau = 0.2$ from the Lemma statement, and the third step follows from simple algebra.

Since $m \leq \delta/\poly(n)$, we are allowed to take union bound (see Fact~\ref{fac:union_bound}) over all $l \in [m]$. 

Thus, we have 
\begin{align*}
F_{\poly}(A; \beta; x; y) 
= & ~ \phi( \sum_{l=0}^m   \phi_{\tau}( \langle f_{\poly}(A; \beta; x) , y_{*,l} \rangle   ) ) \\
= & ~ 0,
\end{align*}
where the first step follows from the definition of $F_{\poly}$ (see Definition~\ref{def:F}) and the second step follows from Eq.~\eqref{eq:phi_tau_0_2}.

{\bf Proof of Part 2.}
Note that all the $y_{*,l}$ are independent, for each $l\in [m]$, we call Part 1 of Lemma~\ref{lem:property_of_dataset_1}, we can show that, with probability $1 - \delta / \poly(n)$,
\begin{align}\label{eq:phi_rau_0}
  \phi_{\tau}( \langle f_{\poly}(A; \beta; x) , y_{*,l} \rangle   ) 
  = & ~ \max \{ \langle f_{\poly}(A; \beta; x) , y_{*,l} \rangle - \tau, 0  \} \notag\\
  \leq & ~ \max \{ C \frac{ \sqrt{ \log(n/\delta)} }{\sqrt{ n } } \cdot 16^{\beta} - \tau, 0  \} \notag\\
  \leq & ~ \max \{ 0.1 - \tau, 0  \} \notag\\
  \leq & ~ \max \{ 0.1 - 0.2, 0  \} \notag\\
  = & ~ 0,
\end{align}
where the first step follows from the definition of $\phi_{\tau}$ (see Definition~\ref{def:relu}), the second step follows from Part 1 of Lemma~\ref{lem:property_of_dataset_1}, the third step follows from $\log(n) = O(n)$, the fourth step follows from $\tau = 0.2$ (see the Lemma statement), and the last step follows from simple algebra.

Since $m \leq \delta/\poly(n)$, we are allowed to take union bound (see Fact~\ref{fac:union_bound}) over all $l \in [m]$.

Thus, we have 
\begin{align*}
F_{\poly}(A; \beta; x; y) 
= & ~ \phi( \sum_{l=0}^m   \phi_{\tau}( \langle f_{\poly}(A; \beta; x) , y_{*,l} \rangle   ) ) \\
= & ~ 0,
\end{align*}
where the first step follows from the definition of $F_{\poly}$ (see Definition~\ref{def:F}) and the second step follows from Eq.~\eqref{eq:phi_rau_0}.
\end{proof}

\section{Self-Attention Dataset}
\label{sec:app_self_attention}

In Section \ref{sec:self_dataset}, we give the definition of the dataset used in the following sections for self-attention.
In Section~\ref{sub:app_self_attention:functions}, we define more important functions. 
In Section~\ref{sub:app_self_attention:main}, we present the main result of this paper, namely the summary of the key properties presented in later sections.
In Section \ref{sec:app:data_1_with_exp}, we show that the output of the $F_{\poly}$ is greater than $0$ with high probability.
In Section \ref{sec:app:data_0_with_exp}, we show that the output of $F_{\poly}$ is equal to $0$ with high probability. 
In Section \ref{sec:app:data_1_with_lin}, we show that the output of the $F_{\poly}$ is equal to $0$ with high probability.
In Section \ref{sec:app:data_0_with_lin}, we show that the output of $F_{\poly}$ is equal to $0$ with high probability.

\subsection{Definition of Dataset}
\label{sec:self_dataset} 

In this section, we give the formal definitions of the dataset.

\begin{definition}[Self-Attention dataset distribution]\label{def:dataset_self_attention}
We define
\begin{itemize}
    \item $a_0 \in (0, 0.1)$
    \item Let $a_1 \geq 0.7$
    \item $b + c = 1$ for $b \geq 0.1 , c \geq 0.1$
\end{itemize}

Given two sets of datasets ${\cal D}_0$ and ${\cal D}_1$ 
\begin{itemize}
    \item For each $\{ A_1, A_2, A_3 \} \in {\cal D}_0$, we have
    \begin{itemize}
        \item $A_1 = A_2 = A_3 \in \R^{n \times d}$ 
        \item Assume $n = (d-2) t$ where $t$ is a positive integer
        \item Type I column: the first column of $A_1$ is $e_{j_3} \cdot a_0 $ for some $j_3 \in [n]$  
        \item Type II column: from the second column to the $(d-1)$-th columns of $A_1$ are $\begin{bmatrix} I_{d-2} \\ I_{d-2} \\ \vdots \\ I_{d-2} \end{bmatrix} \cdot b  $
        \item Type III column: the last column of $A_1$ is ${\bf 1}_n \cdot c $
    \end{itemize}
    \item For each $\{ A_1, A_2, A_3 \} \in {\cal D}_1$, we have
    \begin{itemize}
        \item $A_1 = A_2 = A_3 \in \R^{n \times d}$
        \item Assume $n = (d-2) t$ where $t$ is a positive integer
        \item Type I column: the first column of $A_1$ is $e_{j_3} \cdot a_1  $ for $j_3 \in [n]$
        \item Type II column: from the second column to the $(d-1)$-th columns of $A_1$ are  
        $\begin{bmatrix} I_{d-2} \\ I_{d-2} \\ \vdots \\ I_{d-2} \end{bmatrix} \cdot b  $
        \item Type III column: the last column of $A_1$ is ${\bf 1}_n \cdot c $
    \end{itemize}
\end{itemize}
\end{definition}

\begin{figure}[!ht]
    \centering
    \includegraphics[width = 0.8\linewidth]{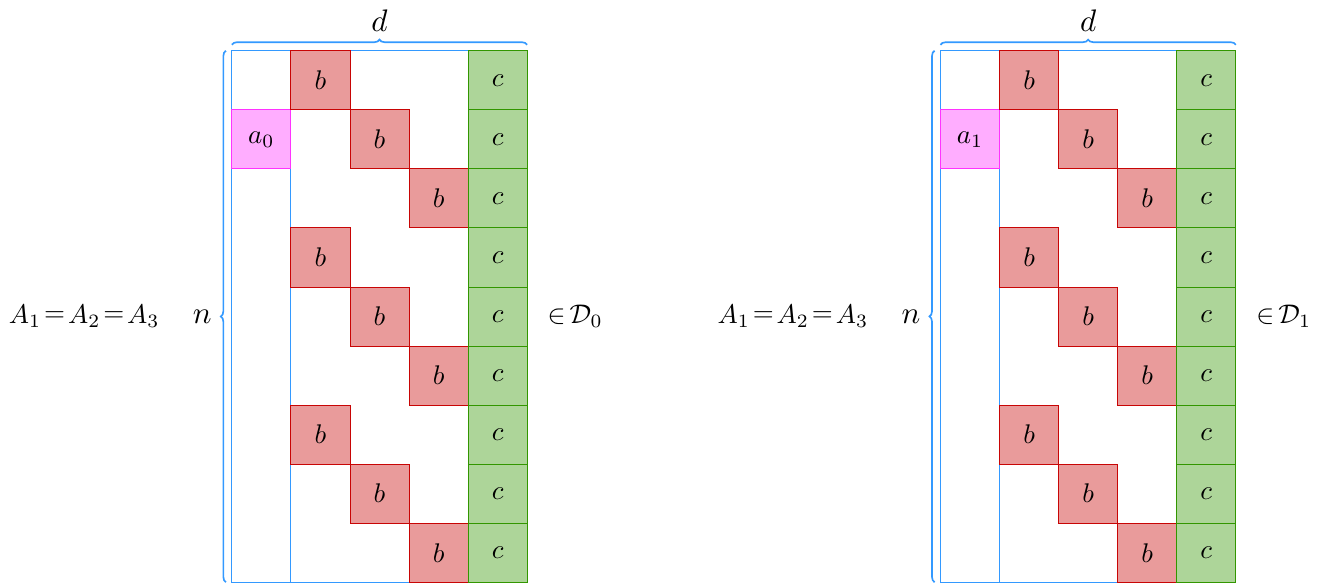}
    \caption{The visualization of the matrices $A_1, A_2, A_3$ (see Definition~\ref{def:dataset_self_attention}). Here, we use $A_1 = A_2 = A_3 \in \mathcal{D}_0$ and $A_1 = A_2 = A_3 \in \mathcal{D}_1$ as an abbreviation of $A_1 = A_2 = A_3 \in \{A_1, A_2, A_3\} \in \mathcal{D}_0$ and $A_1 = A_2 = A_3 \in \{A_1, A_2, A_3\} \in \mathcal{D}_1$. In this figure, we take $n = 9$, $d = 5$, $j_3 = 2$, and $t = 3$. The left column is the type I column, which contains the pink and white colors. The middle column is the type II column, which contains the red and white colors. The right column is the type III column, which contains the green color. The square of white color represents the $0$ entry, the square of red color represents the entry of $b \in \R$, the square of green color represents the entry of $c \in \R$, and the square of pink color represents the entry of $a_0, a_1 \in \R$ ($a_0$ is the entry of the matrices in $\mathcal{D}_0$, namely the figure on the left; $a_1$ is the entry of the matrices in $\mathcal{D}_1$, namely the figure on the right).}
    \label{fig:dataset}
\end{figure}

\subsection{Definitions of Functions}
\label{sub:app_self_attention:functions}

In this section, we define more functions.

\begin{definition}[Polynomial Attention]\label{def:poly_attention}
Let $A_1, A_2, A_3 \in \R^{n \times d}$.

Let $Q, K \in \R^{d \times d}$ (see Definition~\ref{def:attention_poly_unit}).

Let $x = \vect(QK^\top) \in \R^{d^2}$. 

Let $\A = A_1 \otimes A_2 \in \R^{n^2 \times d^2}$. 

Let $\A_{j_0} \in \R^{n \times d^2}$ denote the $j_0$-th block of $\A$.

Let $g : \R \rightarrow \R$ be defined in Definition~\ref{def:attention_poly_unit}.

Let $V \in \R^{d \times d}$.

For each $j_0 \in [n]$, we define $u_{\poly}(\A,x)_{j_0} \in \R^{n}$ as follows
\begin{align*}
    u_{\poly}(\A,x)_{j_0} := g( \A_{j_0} x )
\end{align*}
(Here we apply $g$ entry-wisely, i.e., $g(\A_{j_0}x)_{j_1} = g( (\A_{j_0} x)_{j_1} )$ for all $j_1 \in [n]$)

We define $\alpha_{\poly}(\A,x)_{j_0}$ as follows
\begin{align*}
 \alpha_{\poly}(\A,x)_{j_0} := \langle  u_{\poly}(\A,x)_{j_0}, {\bf 1}_n \rangle
\end{align*}
We define $f_{\poly}(\A,x)_{j_0}$ as follows
\begin{align*}
    f_{\poly}(\A,x)_{j_0} = \alpha_{\poly}(\A,x)_{j_0}^{-1} \cdot u_{\poly}(\A,x)_{j_0}
\end{align*}
We define $c_{\poly}(\A,x)_{j_0} \in \R^d$ as follows
\begin{align*}
    c_{\poly}(\A,x)_{j_0,i_0}:= \langle f_{\poly}(\A,x)_{j_0} , (A_3 V)_{i_0} \rangle, ~~~\forall i_0 \in [d]
\end{align*}

\end{definition}

We state an equivalence formula which will be implicitly used in many proofs.
\begin{claim}[Equivalence Formula]\label{cla:equivalence_formula}

If the following conditions hold,
\begin{itemize}
    \item Let $Q, K \in \R^{d \times d}$ (see Definition~\ref{def:attention_poly_unit}).
    \item Let $x = \vect(QK^\top) \in \R^{d^2}$
    \item Let $(A_1)_{j_0,*} \in \R^d$ denote the $j_0$-th row of $A_1 \in \R^{n \times d}$
    \item Let $A_2 \in \R^{n \times d}$
    \item Let $\A = A_1 \otimes A_2 \in \R^{n^2 \times d^2}$
\end{itemize}
Then, we have
\begin{itemize}
    \item $u_{\poly}(\A,x)_{j_0} =  ( g( (A_1)_{j_0,*} Q K^\top A_2^\top ) )^\top$
    \item $\alpha_{\poly}(\A,x)_{j_0} = \langle  ( g( (A_1)_{j_0,*} Q K^\top A_2^\top ) )^\top, {\bf 1}_n \rangle$ 
\end{itemize}
\end{claim}

\begin{figure}[!ht]
    \centering
    \includegraphics[width = 0.8\linewidth]{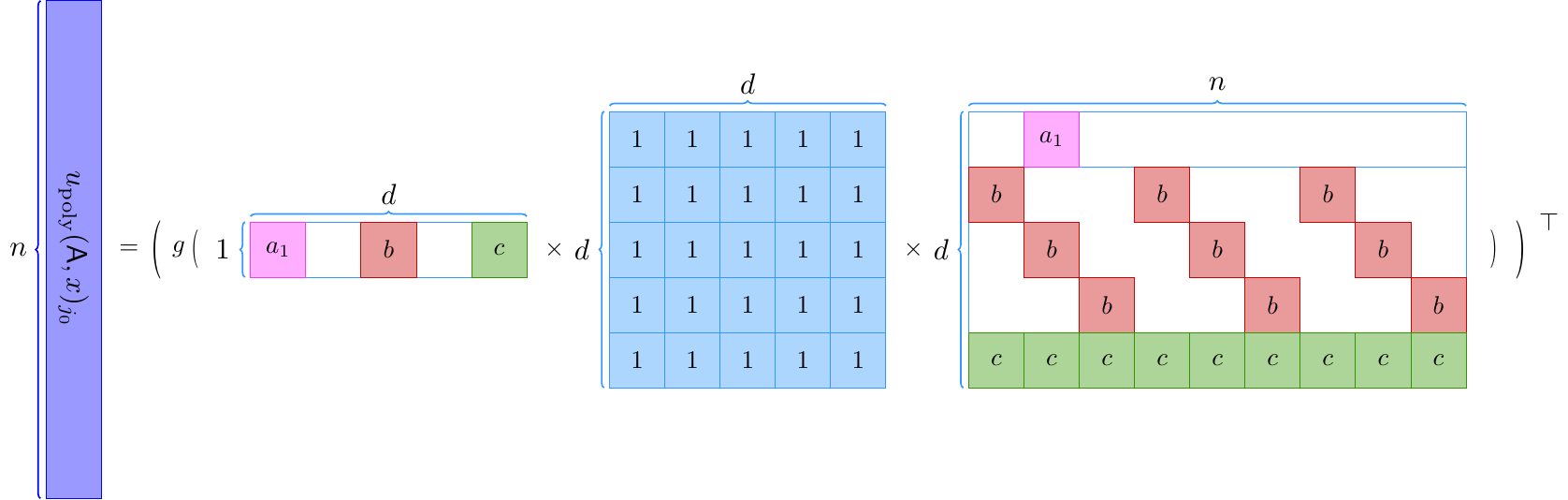}
    \caption{The visualization of $u_{\poly}(\A,x)_{j_0} =  ( g( (A_1)_{j_0,*} Q K^\top A_2^\top ) )^\top$ (see Claim~\ref{cla:equivalence_formula}). In this figure, we take $n = 9$, $d = 5$, $j_3 = j_0 = 2$, and $t = 3$. We let $A_1, A_2 \in \mathcal{D}_1$. The square of white color represents the $0$ entry, the square of red color represents the entry of $b \in \R$, the square of light blue color represents the entry of $1$, the square of green color represents the entry of $c \in \R$, and the square of pink color represents the entry of $a_1 \in \R$.}
    \label{fig:equivalence_formula_eq_1}
\end{figure}

\begin{figure}[!ht]
    \centering
    \includegraphics[width = 0.8\linewidth]{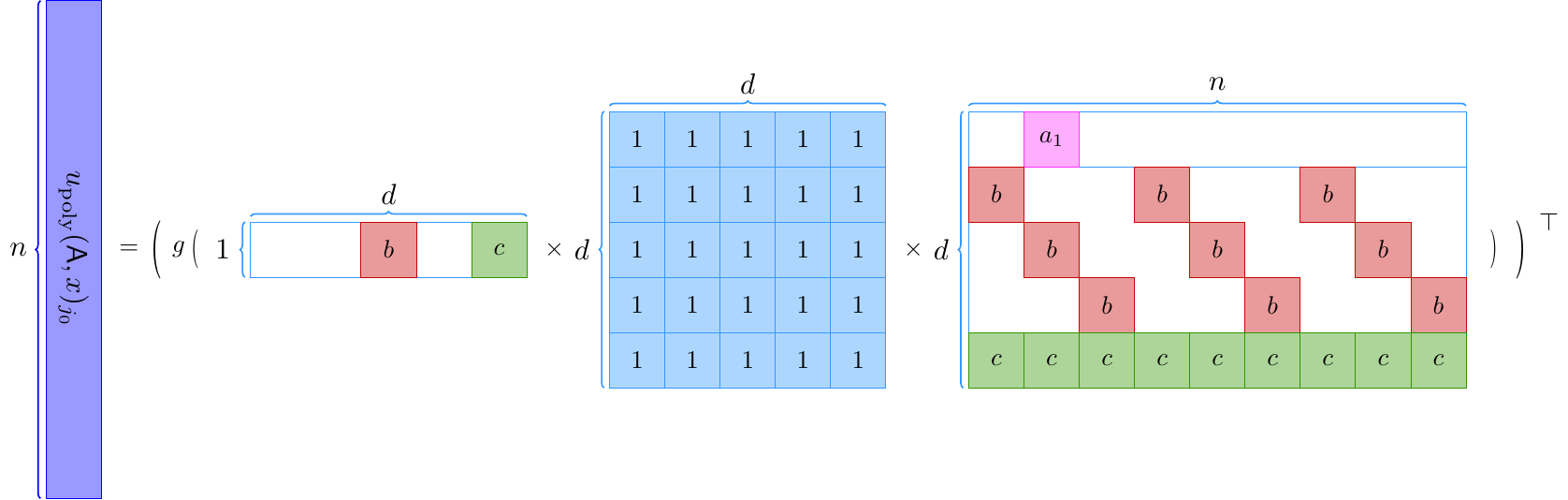}
    \caption{The visualization of $u_{\poly}(\A,x)_{j_0} =  ( g( (A_1)_{j_0,*} Q K^\top A_2^\top ) )^\top$ (see Claim~\ref{cla:equivalence_formula}). In this figure, we take $n = 9$, $d = 5$, $j_3 = 2 \neq 5 = j_0$, and $t = 3$. We let $A_1, A_2 \in \mathcal{D}_1$. The square of white color represents the $0$ entry, the square of red color represents the entry of $b \in \R$, the square of light blue color represents the entry of $1$, the square of green color represents the entry of $c \in \R$, and the square of pink color represents the entry of $a_1 \in \R$.}
    \label{fig:equivalence_formula_eq_12}
\end{figure}

\begin{figure}[!ht]
    \centering
    \includegraphics[width = 0.7\linewidth]{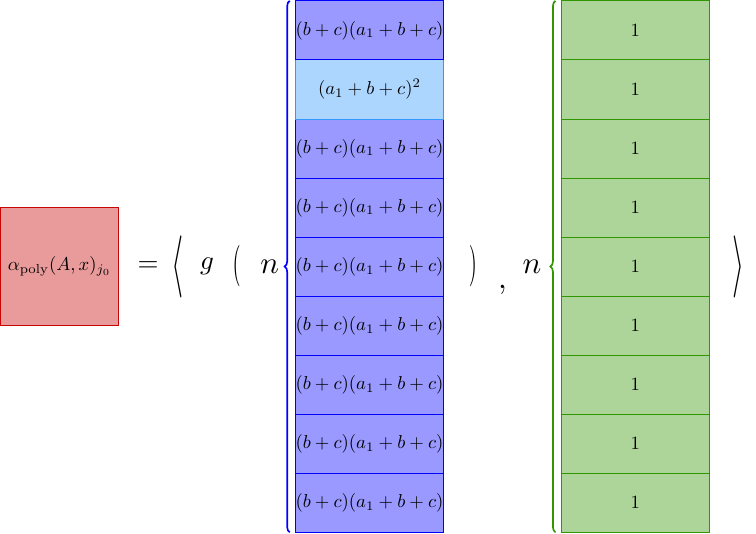}
    \caption{The visualization of $\alpha_{\poly}(\A,x)_{j_0} = \langle  ( g( (A_1)_{j_0,*} Q K^\top A_2^\top ) )^\top, {\bf 1}_n \rangle$ (see Claim~\ref{cla:equivalence_formula}), where the computation of $( g( (A_1)_{j_0,*} Q K^\top A_2^\top ) )^\top$ (the blue vector) can be visualized in Figure~\ref{fig:equivalence_formula_eq_1}, namely $j_3 = j_0$. We let $A_1, A_2 \in \mathcal{D}_1$. The dark blue rectangles represent $(b + c)(a_1 + b + c) \in \R$, the light blue rectangles represent $(a_1 + b + c)^2 \in \R$, and the green rectangles represent $1$. The red square presents the real number $\alpha_{\poly}(\A,x)_{j_0}$.}
    \label{fig:equivalence_formula_eq_2}
\end{figure}

\begin{figure}[!ht]
    \centering
    \includegraphics[width = 0.7\linewidth]{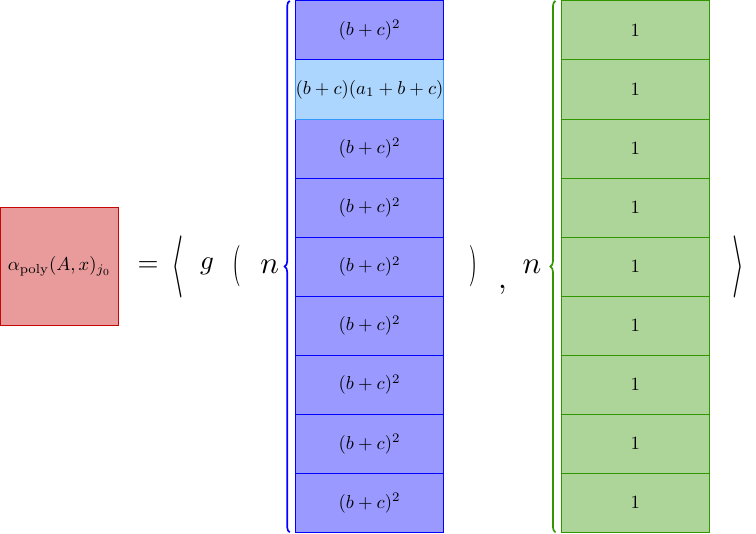}
    \caption{The visualization of $\alpha_{\poly}(\A,x)_{j_0} = \langle  ( g( (A_1)_{j_0,*} Q K^\top A_2^\top ) )^\top, {\bf 1}_n \rangle$ (see Claim~\ref{cla:equivalence_formula}), where the computation of $( g( (A_1)_{j_0,*} Q K^\top A_2^\top ) )^\top$ (the blue vector) can be visualized in Figure~\ref{fig:equivalence_formula_eq_12}, namely $j_3 \neq j_0$. We let $A_1, A_2 \in \mathcal{D}_1$. The dark blue rectangles represent $(b + c)^2 \in \R$, the light blue rectangles represent $(b + c)(a_1 + b + c) \in \R$, and the green rectangles represent $1$. The red square presents the real number $\alpha_{\poly}(\A,x)_{j_0}$.}
    \label{fig:equivalence_formula_eq_22}
\end{figure}

\begin{figure}[!ht]
    \centering
    \includegraphics[width = 0.8\linewidth]{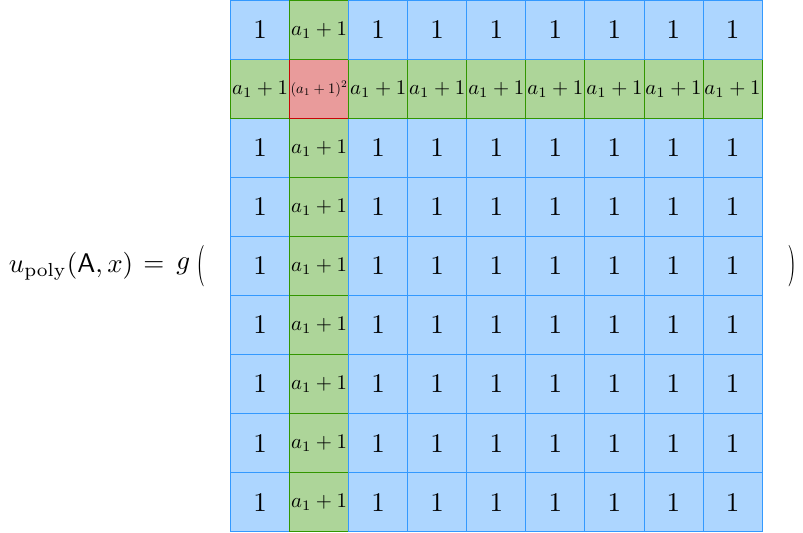}
    \caption{The visualization of $u_{\poly}(\A,x) \in \R^{n \times n}$. In this figure, we take $n = 9$. The 1st and the 3rd to 9th columns are taken from Figure~\ref{fig:equivalence_formula_eq_12} and the second column is taken from Figure~\ref{fig:equivalence_formula_eq_1}. We use the fact that $b + c = 1$ to simplify the algebraic expression. For specific computation details, one can refer to Lemma~\ref{lem:self_dataset_1_f_exp}.}
    \label{fig:u_poly}
\end{figure}

\begin{figure}[!ht]
    \centering
    \includegraphics[width = 0.8 \linewidth]{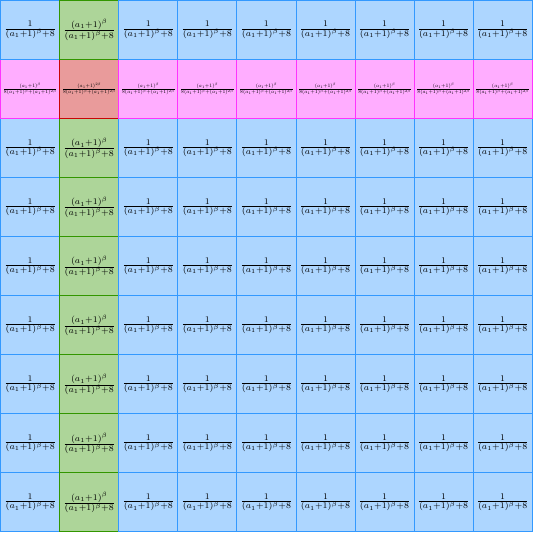}
    \caption{The visualization of $f_{\poly}(\A,x) \in \R^{n \times n}$. In this figure, we take $n = 9$, $d = 5$, $j_3 = 2$, and $t = 3$. We let $A_1, A_2 \in \mathcal{D}_1$. This figure is to normalize each entry of $u_{\poly}(\A,x)$ (see Figure~\ref{fig:u_poly}) based on the definition of $f_{\poly}(\A,x)$ (see Definition~\ref{def:poly_attention}). The blue squares represent $\frac{1}{(\alpha_1 + 1)^\beta + 8} \in \R$, the green squares represent $\frac{(\alpha_1 + 1)^\beta}{(\alpha_1 + 1)^\beta + 8} \in \R$, the pink squares represent $\frac{(a_1 + 1)^\beta}{8 (a_1 + 1)^\beta + (a_1 + 1)^{2\beta}} \in \R$, and the red square represents $\frac{(a_1 + 1)^{2\beta}}{8 (a_1 + 1)^\beta + (a_1 + 1)^{2\beta}} \in \R$.}
    \label{fig:f_poly}
\end{figure}

\begin{figure}[!ht]
    \centering
    \includegraphics[width = 0.7 \linewidth]{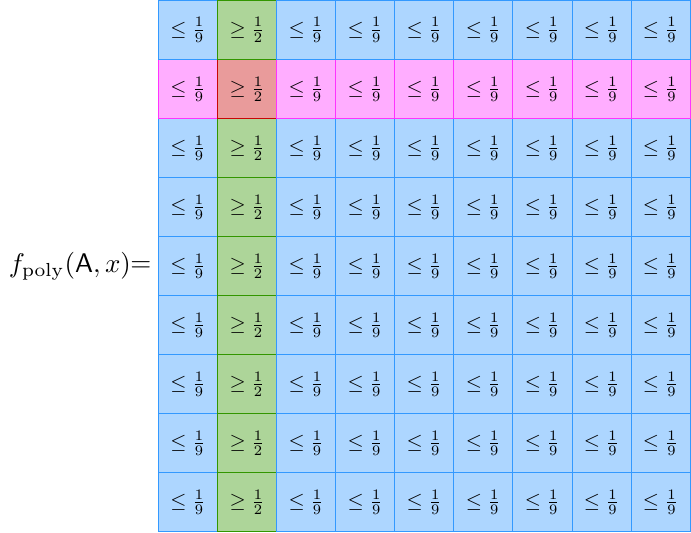}
    \caption{The simplification of Figure~\ref{fig:f_poly}. In this figure, we take $n = 9$, $d = 5$, $j_3 = 2$, and $t = 3$. We let $A_1, A_2 \in \mathcal{D}_1$. By simple algebra, we can get that the blue squares in Figure~\ref{fig:f_poly} $\frac{1}{(\alpha_1 + 1)^\beta + 8} \in \R$ are less than $1/9$, the green squares in Figure~\ref{fig:f_poly} $\frac{(\alpha_1 + 1)^\beta}{(\alpha_1 + 1)^\beta + 8} \in \R$ are greater than $1/2$, the pink squares in Figure~\ref{fig:f_poly} $\frac{(a_1 + 1)^\beta}{8 (a_1 + 1)^\beta + (a_1 + 1)^{2\beta}} \in \R$ are less than $1/9$, and the red square in Figure~\ref{fig:f_poly} $\frac{(a_1 + 1)^{2\beta}}{8 (a_1 + 1)^\beta + (a_1 + 1)^{2\beta}} \in \R$ is greater than $1/2$. Moreover, the blue squares are the examples of $j_0 \neq j_3$ and $j_1 \neq j_3$ (see Part 4b of Lemma~\ref{lem:self_dataset_1_f_exp}), the green squares are the examples of $j_0 \neq j_3$ and $j_1 = j_3$ (see Part 4a of Lemma~\ref{lem:self_dataset_1_f_exp}), the pink squares are the examples of $j_0 = j_3$ and $j_1 \neq j_3$ (see Part 3b of Lemma~\ref{lem:self_dataset_1_f_exp}), and the red square is the example of $j_0 = j_3$ and $j_1 = j_3$ (see Part 3a of Lemma~\ref{lem:self_dataset_1_f_exp}).}
    \label{fig:f_poly_2}
\end{figure}

\begin{proof}

{\bf Proof of Part 1}

We have
\begin{align*}
   \underbrace{ u_{\poly}(\A,x)_{j_0} }_{n \times 1}
    = & ~ g( \A_{j_0} \cdot x )\\
    = & ~ g( \A_{j_0} \cdot \vect(QK^\top) )\\
    = & ~ g( (A_1 \otimes A_2)_{j_0} \cdot \vect(QK^\top) )\\
    = & ~ g( ((A_1)_{j_0,*} \otimes A_2) \cdot \vect(QK^\top) )\\
    = & ~ \underbrace{ g( (A_1)_{j_0,*} \cdot QK^\top \cdot A_2^\top )^\top }_{n \times 1}
\end{align*}
where the first step follows from the definition of $u_{\poly}$ (see Definition~\ref{def:poly_attention}), the second step follows from the definition of $x$ (see from the Lemma statement), the third step follows from the definition of $\A$ (see from the Lemma statement), the fourth step follows from the property of Kronecker product, and the last step follows from $\underbrace{ g( (A_1)_{j_0,*} Q K^\top A_2^\top ) }_{1 \times n}$ is a row vector, that's why we need to apply transpose to turn it into a column vector $\underbrace{ g( (A_1)_{j_0,*} \cdot QK^\top \cdot A_2^\top )^\top }_{n \times 1}$.

{\bf Proof of Part 2}

We have
\begin{align*}
    \alpha_{\poly}(\A,x)_{j_0} 
    = & ~ \langle  u_{\poly}(\A,x)_{j_0}, {\bf 1}_n \rangle\\
    = & ~ \langle  ( g( (A_1)_{j_0,*} Q K^\top A_2^\top ) )^\top, {\bf 1}_n \rangle,
\end{align*}
where the first step follows from the definition of $\alpha_{\poly}$ (see Definition~\ref{def:poly_attention}) and the second step follows from {\bf Part 1}.
\end{proof}

\begin{definition}
Let $y\in \R^{n \times m}$. 

Let $y_{*,l} \in \R^n$ denote the $l$-th column of $y$. 

We define
\begin{align*}
F_{\poly}(A_1,A_2,A_3) := \phi( \sum_{j_0=1}^n \sum_{l=1}^m \phi_{\tau} ( \langle c_{\poly} (\A,x)_{j_0}, y_{*,l} \rangle  ) )
\end{align*}
\end{definition}

\subsection{Main Result}
\label{sub:app_self_attention:main}

We state our main result as follows:
\begin{theorem}[Main result, formal version of Theorem~\ref{thm:main:informal}]\label{thm:main:formal}
If the following conditions hold
\begin{itemize}
    \item Let $\tau = c+ 0.1$
    \item Let $m = O( \log (n/\delta) )$
\end{itemize}
Then, we have
\begin{itemize}
        \item For sufficiently large $\beta$
        \begin{itemize}
            \item For any $A_1, A_2, A_3 $ from ${\cal D}_1$, we have $F_{\poly}(A_1,A_2, A_3) > 0$
            \item For any $A_1, A_2, A_3 $ from ${\cal D}_0$, we have $F_{\poly}(A_1,A_2, A_3) = 0$
        \end{itemize}
        \item For sufficiently small $\beta$
        \begin{itemize}
            \item For any $A_1, A_2, A_3 $ from ${\cal D}_1$, we have $F_{\poly}(A_1,A_2, A_3) = 0$
            \item For any $A_1, A_2, A_3 $ from ${\cal D}_0$, we have $F_{\poly}(A_1,A_2, A_3) = 0$
        \end{itemize}
\end{itemize}
\end{theorem}
\begin{proof}
It follows from combining Theorem~\ref{thm:main:formal:part1}, \ref{thm:main:formal:part2}, \ref{thm:main:formal:part3}, and \ref{thm:main:formal:part4}.
\end{proof}

\subsection{Dataset 1 When Applying \texorpdfstring{$F_{\poly}$}{} With Large \texorpdfstring{$\beta$}{}--A High-Degree Polynomial Case}\label{sec:app:data_1_with_exp}
In Section~\ref{sec:ds_1_u_f_exp} we analyze the property of dataset 1 with respect to function $u_{\poly}$ and $f_{\poly}$. In Section~\ref{sec:ds_1_c_exp} we analyze the property of dataset 1 with respect to function $c_{\poly}$. In Section~\ref{sec:ds_1_y_exp} we analyze the property of dataset 1 with respect to function $c_{\poly}$ with random signs. In Section~\ref{sec:ds_1_F_exp} we show the property of dataset 1 with respect to the output of $F_{\poly}$. 

\subsubsection{The Property of Dataset 1 When Applying Functions \texorpdfstring{$u_{\poly}$}{} and \texorpdfstring{$f_{\poly}$}{}}
\label{sec:ds_1_u_f_exp}

In this section, we analyze the key properties of Dataset 1 while applying functions $u_{\poly}$ and $f_{\poly}$.

\begin{lemma}\label{lem:self_dataset_1_f_exp}
If the following conditions hold
\begin{itemize} 
    \item Let $\beta \geq \log n$
    \item Let $a_1 \geq 1$
    \item Let $(a_1+1)^{\beta} \geq n$
    \item Let $b+c = 1$
    \item Let $\{ A_1, A_2, A_3 \}$ from dataset ${\cal D}_1$ (see Definition~\ref{def:dataset_self_attention})
    \item Let $QK^\top = {\bf 1}_{d \times d}$  
\end{itemize} 
Then, for $u_{\poly}(\A,x)_{j_0,j_1}$ and $f_{\poly}(\A,x)_{j_0,j_1}$ entry we have
\begin{itemize}
    \item {\bf Part 1.} For $j_0 = j_3$, 
    \begin{itemize}
        \item {\bf Part 1a.} For $j_1 = j_3$, then  $u_{\poly}(\A,x)_{j_0,j_1} = (a_1+1)^{2\beta}$.
        \item {\bf Part 1b.} For $j_1 \neq j_3$, then  $u_{\poly}(\A,x)_{j_0,j_1} = (a_1+1)^{\beta} $.
    \end{itemize}
    \item {\bf Part 2.} For $j_0 \neq j_3$
    \begin{itemize}
        \item {\bf Part 2a.} For $j_1 = j_3$, then $u_{\poly}(\A,x)_{j_0,j_1} = (a_1+1)^{\beta} $.
        \item {\bf Part 2b.} For $j_1 \neq j_3$, then $u_{\poly}(A,x)_{j_0,j_1} = 1$.
    \end{itemize}
    \item {\bf Part 3.} For $j_0 = j_3$,
    \begin{itemize}
        \item {\bf Part 3a.} For $j_1 = j_3$, then  $f_{\poly}(\A,x)_{j_0,j_1} \geq 1/2$.  
        \item {\bf Part 3b.} For $j_1 \neq j_3$, then  $f_{\poly}(\A,x)_{j_0,j_1} \leq 1/n$.
    \end{itemize}
    \item {\bf Part 4.} For $j_0 \neq j_3$,
    \begin{itemize}
        \item {\bf Part 4a.} For $j_1 = j_3$, then  $f_{\poly}(\A,x)_{j_0,j_1} \geq 1/2 $. 
        \item {\bf Part 4b.} For $j_1 \neq j_3$, then  $f_{\poly}(\A,x)_{j_0,j_1} \leq 1/n$. 
    \end{itemize}
\end{itemize}
\end{lemma}
\begin{proof}
 
{\bf Proof of Part 1.}

{\bf Proof of Part 1a.}

For $j_0 = j_3$ and $j_1 = j_3$, by computing the tensor product of two $3$-sparse vector
\begin{align*}
& ~ u_{\poly}(\A,x)_{j_0,j_1} \notag \\
= & ~ g(  \langle  (a_1 , 0 , \cdots, 0, b  , 0, \cdots, 0, c  ) \otimes (a_1 , 0 , \cdots, 0, b  , 0, \cdots, 0, c  ) , {\bf 1}_{d \times d} \rangle ) \\ 
= & ~ g( (a_1 + b + c)^2   )   \\
= & ~ (a_1+b+c)^{2\beta}\\
= & ~ (a_1+1)^{2\beta}
\end{align*}
where the first step follows from Definition \ref{def:dataset_self_attention} and Definition \ref{def:poly_attention}, the second step follows from algebra, the second step follows from simple algebra, the third step follows from simple algebra, and the last step follows from $b + c = 1$.

{\bf Proof of Part 1b.}
For $j_0 = j_3$ and $j_1 \neq j_3$. 
\begin{align*}
u_{\poly}(\A,x)_{j_0,j_1} 
= & ~ g( (a_1   + b  + c   ) ( b   + c  )  ) \\
= & ~ g(  (b + c)^2 + a_1(b+c)  ) \\
= & ~ g(  1 + a_1  ) \\
= & ~ (  1 + a_1  )^\beta
\end{align*}
where the first step follows from Definition \ref{def:dataset_self_attention} and Definition \ref{def:poly_attention},  
the second step follows from simple algebra, the third step follows from $b + c = 1$ (see from the Lemma statement), and the last step follows from the definition of $g$ (see Definition~\ref{def:attention_poly_unit}).

{\bf Proof of Part 2.}

{\bf Proof of Part 2a.}

For $j_0 \neq j_3$ and $j_1 = j_3$, this case is the same as Part 1b.

{\bf Proof of Part 2b.}

For $j_0 \neq j_3$ and $j_1 \neq j_3$,

For the case that we tensor a $2$-sparse vector with another $2$-sparse vector, we have 
\begin{align*}
u_{\poly}(\A,x)_{j_0,j_1} 
= & ~ g( ( b  + c  )^2 ) \\
= & ~ ( b + c )^{2 \beta}\\
= & ~ 1^{2 \beta}\\
= & ~ 1,
\end{align*}
where the first step follows from Definition \ref{def:dataset_self_attention} and Definition \ref{def:poly_attention}, the second step follows from the definition of $g$ (see Definition~\ref{def:attention_poly_unit}), the third step follows from $b + c = 1$ (see from the Lemma statement), and the last step follows from simple algebra.

{\bf Proof of Part 3.}

{\bf Proof of Part 3a.}

We can show
\begin{align*}
f_{\poly}(\A,x)_{j_0,j_1} 
= & ~ \langle  u_{\poly}(\A,x)_{j_0}, {\bf 1}_n \rangle^{-1} \cdot u_{\poly}(\A,x)_{j_0,j_1}\\ 
= & ~ \frac{1}{ (a_1+b+c)^{2\beta} + (n-1) \cdot ((a_1+b+c) (b+c))^{\beta} } \cdot (a_1+b+c)^{2\beta}  \\
\geq & ~ \frac{1}{2}
\end{align*}
where the first step follows from the definition of $f_{\poly}$, the second step follows from  {\bf Part 1a} and {\bf Part 1b}, and the last step follows from $(a_1+b+c)^{2\beta} > n (a_1+b+c)^{\beta} (b+c)^{\beta}$ (which is equivalent to $(a_1+b+c)^{\beta} > n (b+c)^{\beta}$).

{\bf Proof of Part 3b.}
\begin{align*}
f_{\poly}(\A,x)_{j_0,j_1} 
= & ~ \langle  u_{\poly}(\A,x)_{j_0}, {\bf 1}_n \rangle^{-1} \cdot u_{\poly}(\A,x)_{j_0,j_1}\\ 
= & ~ \frac{1}{ (a_1+b+c)^{2\beta} + (n-1) \cdot( (a_1+b+c)(b+c) )^{\beta} } \cdot  ( (a_1+b+c)(b+c) )^{\beta}  \\
\leq & ~ \frac{1}{ n \cdot( (a_1+b+c)(b+c) )^{\beta} } \cdot  ( (a_1+b+c)(b+c) )^{\beta}  \\
= & ~ 1/n
\end{align*}
where the first step follows from the definition of $f_{\poly}$, the second step follows from {\bf Part 1a} and {\bf Part 1b}, the third step follows from $( (a_1+b+c)(b+c) )^{\beta} \leq (a_1+b+c)^{2\beta}$, and the last step follows from simple algebra.

{\bf Proof of Part 4.}

{\bf Proof of Part 4a.}

We can show
\begin{align*}
f_{\poly}(\A,x)_{j_0,j_1} 
= & ~ \langle  u_{\poly}(\A,x)_{j_0}, {\bf 1}_n \rangle^{-1} \cdot u_{\poly}(\A,x)_{j_0,j_1}\\ 
= & ~ \frac{1}{ ( (a_1+b+c)(b+c) )^{\beta} + (n-1) \cdot (b+c)^{2\beta} } \cdot ( (a_1+b+c)(b+c) )^{\beta} \\
\geq  & ~ \frac{1}{2}
\end{align*}
where the first step follows from the definition of $f_{\poly}$, the second step follows from  {\bf Part 2a} and {\bf Part 2b},
and the last step follows from simple algebra.

{\bf Proof of Part 4b.}

We can show
\begin{align*}
f_{\poly}(\A,x)_{j_0,j_1} 
= & ~ \langle  u_{\poly}(\A,x)_{j_0}, {\bf 1}_n \rangle^{-1} \cdot u_{\poly}(\A,x)_{j_0,j_1}\\ 
= & ~ \frac{1}{ ( (a_1+b+c)(b+c) )^{\beta} + (n-1) \cdot (b+c)^{2\beta} } \cdot (b+c) ^{2\beta}  \\
\leq & ~ \frac{1}{n}
\end{align*}
where the first step follows from the definition of $f_{\poly}$, the second step follows from {\bf Part 2a} and {\bf Part 2b}, and the last step follows from simple algebra.
\end{proof}

\subsubsection{The Property of Dataset 1 When Applying the Function \texorpdfstring{$c_{\poly}$}{}}
\label{sec:ds_1_c_exp}

In this section, we analyze the key properties of Dataset 1 while applying the function $c_{\poly}$.

\begin{lemma}\label{lem:self_dataset_1_c_exp}
If the following conditions hold
\begin{itemize}
    \item Let $\{A_1, A_2, A_3 \}$ from dataset ${\cal D}_1$ (see Definition~\ref{def:dataset_self_attention})
    \item Let $QK^\top = {\bf 1}_{d \times d}$
    \item Let $V = I_d$
    \item Let $n = t(d-2)$
\end{itemize}
Then for $c_{\poly}(\A,x)_{j_0,i_0} $ entry we have
\begin{itemize}
    \item {\bf Part 1.} For $j_0=j_3$ and $i_0 = 1$, we have $ c_{\poly}(\A,x)_{j_0,i_0}  \geq \frac{1}{2} \cdot a_1  $
    \item {\bf Part 2.} For $j_0=j_3$ and $i_0 \in \{2,\cdots,d-1\}$
    \begin{itemize}
        \item There is only one $i_0$, we have $c_{\poly}(\A,x)_{j_0,i_0} \geq \frac{1}{2} b  $
        \item For the rest of $i_0$, we have $c_{\poly}(\A,x)_{j_0,i_0} \leq \frac{t}{n} b   $
    \end{itemize}
    \item {\bf Part 3.} For $j_0 = j_3$ and $i_0 = d$, we have $ c_{\poly}(\A,x)_{j_0,i_0} = c  $ 
    \item {\bf Part 4.} For $j_0 \neq j_3$ and $i_0 = 1$, we have $ c_{\poly}(\A,x)_{j_0,i_0} \geq \frac{1}{2} a_1  $
    \item {\bf Part 5.} For $j_0\neq j_3$ and $i_0 \in \{2,\cdots,d-1\}$, 
    \begin{itemize}
        \item There is only one $i_0$, we have $c_{\poly}(\A,x)_{j_0,i_0} \geq \frac{1}{2} b  $
        \item For the rest of $i_0$, we have $c_{\poly}(\A,x)_{j_0,i_0} \leq \frac{t}{n} b   $
    \end{itemize}
    \item {\bf Part 6.} For $j_0\neq j_3$ and $i_0 = d$, we have $ c_{\poly}(\A,x)_{j_0,i_0} = c  $
\end{itemize}
\end{lemma}
\begin{proof}

{\bf Proof of Part 1.}

It follows from Part 3 of Lemma~\ref{lem:self_dataset_1_f_exp} and Type I column in $A_3$ (Definition~\ref{def:dataset_self_attention}).

We can show for one-hot vector $e_{j_3} \in \R^n$, 
\begin{align*}
\langle f_{\poly}(\A,x)_{j_0}, (A_3 V)_{i_0} \rangle
= & ~ \langle f_{\poly}(\A,x)_{j_0} , e_{j_3} \cdot a_1  \rangle \\
\geq & ~ \frac{1}{2} a_1  
\end{align*}
where the first step follows from Type I column in $A_3$  (Definition~\ref{def:dataset_self_attention}) and the last step follows from Part 3 of Lemma~\ref{lem:self_dataset_1_f_exp}.

{\bf Proof of Part 2.}

It follows from Part 3 of Lemma~\ref{lem:self_dataset_1_f_exp} and Type II column in $A_3$ (Definition~\ref{def:dataset_self_attention}).

There are two cases we need to consider, there is one $(A_3 V)_{i_0}$'s $j_3$ coordinate is $1$, in this situation, we know
\begin{align*}
\langle f_{\poly}(\A, x)_{j_0}, (A_3 V)_{i_0} \rangle \geq \frac{1}{2} \cdot b 
\end{align*}

For the other case, $(A_3 V)_{i_0}$'s $j_3$ coordinate is $0$, in this case we use the property that $(A_3 V)_{i_0}$ is $t$-sparse vector. 

Then, we have
\begin{align*}
\langle f_{\poly}(\A, x)_{j_0}, (A_3 V)_{i_0} \rangle
\leq & ~ t \cdot \frac{1}{n} \cdot b   \\
\leq & ~ \frac{t}{n} \cdot b ,
\end{align*}
where the first step follows from entries in $f_{\poly}(\mathsf{A}, x)_{j_0}$ are bounded by $\frac{1}{n}$ and at most $t$ entries in $(A_3V)_{i_0}$ are $b $ and rest are zeros, the second step follows from simple algebra. 

{\bf Proof of Part 3.}

It follows from the fact that $f_{\poly}(\A,x)_{j_0}$ is a normalized vector, so 
\begin{align}\label{eq:normalized_vector}
    \langle f_{\poly}(\A,x)_{j_0}, {\bf 1}_n\rangle = 1
\end{align}
and Type III column in $A_3$ (Definition~\ref{def:dataset_self_attention}).

We can show
\begin{align*}
\langle f_{\poly}(\A,x)_{j_0}, (A_3 V)_{i_0} \rangle = & ~ \langle f_{\poly}(\A,x)_{j_0}, c  \cdot {\bf 1}_n \rangle \\
= & ~ \langle f_{\poly}(\A,x)_{j_0}, {\bf 1}_n \rangle \cdot c   \\
= & ~ c,
\end{align*}
where the first step follows from definition of $A_3$ and $V= I_d$, the second step follows from Fact~\ref{fac:vector}, and the third step follows from Eq.~\eqref{eq:normalized_vector}.

{\bf Proof of Part 4, 5 and 6.}

They are the same as Part 1,2,3.
\end{proof}

\subsubsection{The Property of Dataset 1 When Applying the Function \texorpdfstring{$c_{\poly}$}{} With Random Signs}
\label{sec:ds_1_y_exp}

In this section, we analyze the key properties of Dataset 1 while applying the function $c_{\poly}$ with random signs.

\begin{lemma}\label{lem:self_dataset_1_random_exp}
If the following conditions hold
\begin{itemize}
    \item Let $\{A_1, A_2,A_3\}$ be from dataset ${\cal D}_1$ (see Definition~\ref{def:dataset_self_attention})
    \item Let $t \sqrt{d} = o(n^{0.99})$ (since $t(d-2) = n$, then we know $\sqrt{d} = \omega(n^{0.01} )$, this implies $d= \omega(n^{0.02})$).  
\end{itemize}
Then, for each random string $\sigma \in \{-1,+1\}^d$, we have
\begin{itemize}
    \item {\bf Part 1.} If $j_0 = j_3$, then $\Pr[ \langle c_{\poly}(\A,x)_{j_0} , \sigma \rangle \geq (c+0.1)   ] \geq 1/10$
    \item {\bf Part 2.} If $j_0 \neq j_3$, then $\Pr[ \langle c_{\poly}(\A,x)_{j_0} , \sigma \rangle \geq (c+0.1)   ] \geq 1/10$
\end{itemize}
\end{lemma}
\begin{proof}
{\bf Proof of Part 1.}

For $j_0 = j_3$, following  Part 1,2,3 of Lemma~\ref{lem:self_dataset_1_c_exp}, there are three cases for 
\begin{align*}
    c_{\poly}(\A,x)_{j_0,i_0}.
\end{align*}

For $i_0 = 1$, we have 
\begin{align}\label{eq:cpoly_1}
    c_{\poly}(\A,x)_{j_0,i_0} \geq \frac{1}{2} a_1
\end{align}

For $i_0 \in \{ 2, \cdots, d-1 \}$, there is one $i_0$ such that 
\begin{align}\label{eq:cpoly_2}
    c_{\poly}(\A,x)_{j_0,i_0} \geq \frac{1}{2} b.
\end{align}

For the rest of $i_0$, we have 
\begin{align*}
    c_{\poly}(\A,x)_{j_0,i_0} \leq \frac{t}{n } b.
\end{align*}

For $i_0 = d$, we have 
\begin{align}\label{eq:cpoly_3}
    c_{\poly}(\A,x)_{j_0,i_0} = c.
\end{align}

Let $S =\{1, i_0' ,d\}$ denote a set of three indices, and $i_0'$ denote that special index.

By hoeffding inequality (Lemma~\ref{lem:hoeffding_bound}), we know that
\begin{align*}
| \langle c_{\poly}(\A,x)_{j_0,[d] \backslash S} , \sigma \rangle  | \leq & ~ O( \sqrt{\log (n/\delta)} ) \cdot \frac{ t \sqrt{d-3} }{ n } \cdot b  \\
\leq & ~ 0.1  
\end{align*}
with probability at least $1-\delta/\poly(n)$. Here the last step due to $t\sqrt{d} = o(n^{0.99})$ and $\poly(\log n) \leq n^{0.01}$.

Therefore, with probability at least $1/8$, we have 
\begin{align*}
    \langle c_{\poly}(\A,x)_{j_0,S} , \sigma \rangle 
    \geq & ~ \frac{1}{2}a_1   + \frac{1}{2} b   + c   \\
    \geq & ~ (c+0.2),
\end{align*} 
where the first step follows from combining Eq.~\eqref{eq:cpoly_1}, Eq.~\eqref{eq:cpoly_2}, and Eq.~\eqref{eq:cpoly_3} and the second step follows from $a_1 \geq 0.5$.

Since the probability that $\sigma_{i_0} = 1$ for all three cases is $\frac{1}{8}$.

Hence, by combining the above two events, we have
\begin{align*}
    \Pr[ \langle c_{\poly}(\A,x)_{j_0} , \sigma \rangle \geq (c + 0.1)  ] \geq 1/10
\end{align*}

{\bf Proof of Part 2.}

The proof is the same as {\bf Part 1}.
\end{proof}

\subsubsection{The Property of Dataset 1 When Applying the Function \texorpdfstring{$F_{\poly}$}{} }
\label{sec:ds_1_F_exp}

In this section, we analyze the key properties of Dataset 1 while applying the function $F_{\poly}$.

\begin{theorem}\label{thm:main:formal:part1}
If the following conditions hold
\begin{itemize}
    \item Let $d \in [ \omega(n^{0.02}) , n ]$
    \item Let $\tau = (c+0.1) $
    \item Let $m = O(\log (n/\delta))$
    \item For any $\{A_1,A_2,A_3\}$ from ${\cal D}_1$ (Definition~\ref{def:dataset_self_attention})
    \item Let $x = {\bf 1}_{d^2}$
    \item Let $F_{\poly}(A_1,A_2,A_3):=\phi( \sum_{j_0=1}^n \sum_{j_1=1}^m \phi_{\tau}( \langle c_{\poly}(\A,x) , y_{j_1} \rangle) )$
\end{itemize}
Then we have
\begin{itemize}
    \item With high probability $1-\delta/\poly(n)$, $F_{\poly}(A_1,A_2,A_3) > 0$
\end{itemize}
\end{theorem}
\begin{proof}
It follows from using Lemma~\ref{lem:self_dataset_1_random_exp}.
\end{proof}

\subsection{Dataset 0 When Applying \texorpdfstring{$F_{\poly}$}{} With Large \texorpdfstring{$\beta$}{}--A High-Degree Polynomial Case}\label{sec:app:data_0_with_exp}
In Section~\ref{sec:ds_0_u_f_exp} we analyse the property of dataset 0 with respect to function $u_{\poly}$ and $f_{\poly}$. In Section~\ref{sec:ds_0_c_exp} we analyse the property of dataset 0 with respect to function $c_{\poly}$. In Section~\ref{sec:ds_0_y_exp} we analyse the property of dataset 0 with respect to function $c_{\poly}$ with random signs. In Section~\ref{sec:ds_0_F_exp} we show the property of dataset 0 with respect to the output of $F_{\poly}$. 

\subsubsection{The Property of Dataset 0 When Applying Functions \texorpdfstring{$u_{\poly}$}{} and \texorpdfstring{$f_{\poly}$}{}}
\label{sec:ds_0_u_f_exp}

In this section, we analyze the key properties of Dataset 0 while applying the functions $u_{\poly}$ and $f_{\poly}$.

\begin{lemma}\label{lem:self_dataset_0_f_exp}
If the following conditions hold
\begin{itemize}
    \item Let $\{ A_1, A_2, A_3 \}$ from dataset ${\cal D}_1$ (see Definition~\ref{def:dataset_self_attention})
    \item Let $QK^\top ={\bf 1}_{d \times d} $
    \item Let $\beta \geq \log n$
    \item Let $a_0 \in (0,0.1)$
    \item Let $b+c = 1$.
    \item Let $(a_0+1)^{\beta} \leq n^{c_0}$ for $c_0 \in (0,0.2)$
\end{itemize} 
Then, for $u_{\poly}(\A,x)_{j_0,j_1}$ and $f_{\poly}(\A,x)_{j_0,j_1}$ entry we have
\begin{itemize}
    \item {\bf Part 1.} For $j_0 = j_3$, 
    \begin{itemize}
        \item {\bf Part 1a.} If $j_1 = j_3$, then  $u_{\poly}(\A,x)_{j_0,j_1} =  (a_0 + 1)^{2 \beta} $.
        \item {\bf Part 1b.} If $j_1 \neq j_3$, then  $u_{\poly}(\A,x)_{j_0,j_1} = (a_0 + 1)^{\beta}$.
    \end{itemize}
    \item {\bf Part 2.} For $j_0 \neq j_3$,
    \begin{itemize}
        \item {\bf Part 2a.} If $j_1 = j_3$, then $u_{\poly}(\A,x)_{j_0,j_1} = (a_0 + 1)^{\beta}$.
        \item {\bf Part 2b.} If $j_1 \neq j_3$, then $u_{\poly}(\A,x)_{j_0,j_1} = 1 $.
    \end{itemize}
    \item {\bf Part 3.} For $j_0 = j_3$, 
    \begin{itemize}
        \item {\bf Part 3a.} If $j_1 = j_3$, then  $f_{\poly}(\A,x)_{j_0,j_1} \leq 1/ n^{1-c_0}$. 
        \item {\bf Part 3b.} If $j_1 \neq j_3$, then  $f_{\poly}(\A,x)_{j_0,j_1} \leq 1/n$.
    \end{itemize}
    \item {\bf Part 4.} For $j_0 \neq j_3$,
    \begin{itemize}
        \item {\bf Part 4a.} If $j_1 = j_3$, then  $f_{\poly}(\A,x)_{j_0,j_1} \leq 1/n^{1-c_0}$.
        \item {\bf Part 4b.} If $j_0 \neq j_3$, then  $f_{\poly}(\A,x)_{j_0,j_1} \leq 1/n$.
    \end{itemize}
\end{itemize}
\end{lemma}
\begin{proof}
 
{\bf Proof of Part 1.}

{\bf Proof of Part 1a.}
For $j_0 = j_3$ and $j_1 = j_3$, by computing the tensor of two $3$-sparse vector, we have
\begin{align*}
u_{\poly}(\A,x)_{j_0,j_1}
= & ~ g( ( a_0  + b  + c  )^2 ) \\
= & ~ ( a_0  + b  + c  )^{2\beta} \\
= & ~ (a_0+1)^{2\beta}  
\end{align*}
where the first step follows from Definition \ref{def:dataset_self_attention} and Definition~\ref{def:poly_attention}, the second step follows from the definition of $g$ (see Definition~\ref{def:attention_poly_unit}), and the last step follows from $b + c = 1$.

{\bf Proof of Part 1b.}
Let $j_0 = j_3$ and $j_1 \neq j_3$.

In this case, we tensor a $2$-sparse vector with another $3$-sparse vector, and we get
\begin{align*}
u_{\poly}(\A,x)_{j_0,j_1} 
= & ~ g( ( b + c   ) \cdot ( a_0   + b    + c  ) ) \\
= & ~ g( 1 \cdot ( a_0   + 1  ) ) \\
= & ~ ( a_0   + 1  )^{\beta},
\end{align*}
where the first step follows from Definition \ref{def:dataset_self_attention} and Definition~\ref{def:poly_attention}, the second step follows from $b + c = 1$, the third step follows from the definition of $g$ (see Definition~\ref{def:attention_poly_unit}).

{\bf Proof of Part 2.}

{\bf Proof of Part 2a.}

For $j_0 \neq j_3$ and $j_1 = j_3$, this case is same as Part 1b.

{\bf Proof of Part 2b.}

For $j_0 \neq j_3$ and $j_1 \neq j_3$,

For the case that we tensor a $2$-sparse vector with another $2$-sparse vector, we have 
\begin{align*}
u_{\poly}(\A,x)_{j_0,j_1} 
= & ~ g( ( b   + c  )^2 ) \\
= & ~ (b+c)^{2\beta}  \\
= & ~ 1^{2\beta}  \\
= & ~ 1,
\end{align*}
where the first step follows from Definition \ref{def:dataset_self_attention} and Definition~\ref{def:poly_attention}, the second step follows from the definition of $g$ (see Definition~\ref{def:attention_poly_unit}), the third step follows from $b+c = 1$, and the last step follows from simple algebra.

{\bf Proof of Part 3.}

{\bf Proof of Part 3a.}
We can show
\begin{align*}
f_{\poly}(\A,x)_{j_0,j_1} = & ~ \langle  u_{\poly}(\A,x)_{j_0}, {\bf 1}_n \rangle^{-1} \cdot u_{\poly}(\A,x)_{j_0,j_1}\\ 
= & ~ \frac{1}{ (a_0+1)^{2\beta} + (n-1) (a_0+1)^{\beta} } \cdot (a_0+1)^{2\beta}\\
\leq & ~ \frac{1}{n (a_0+1)^{\beta}} \cdot (a_0+1)^{2\beta} \\
= & ~ \frac{ (a_0+1)^{\beta} }{n} \\
\leq & ~ \frac{1}{n^{1-c_0}}
\end{align*}
where the first step follows from the definition of $f_{\poly}$, the second step follows from  {\bf Part 1} and {\bf Part 2},  the third step follows from $(a_0+1)^{2\beta} \geq (a_0+1)^{\beta}$, the fourth step follows from simple algebra, and the last step follows from $(a_0+1)^{\beta} \leq n^{c_0}$.

{\bf Proof of Part 3b.}
\begin{align*}
f_{\poly}(\A,x)_{j_0,j_1} 
= & ~ \langle  u_{\poly}(\A,x)_{j_0}, {\bf 1}_n \rangle^{-1} \cdot u_{\poly}(\A,x)_{j_0,j_1}\\ 
= & ~ \frac{1}{ (a_0+1)^{2\beta} + (n-1) (a_0+1)^{\beta} } \cdot (a_0+1)^{\beta}\\
\leq & ~ \frac{1}{n (a_0+1)^{\beta}} \cdot (a_0+1)^{2\beta} \\
\leq & ~ 1/n,
\end{align*}
where the first step follows from the definition of $f_{\poly}$, the second step follows from  {\bf Part 1} and {\bf Part 2},  the third step follows from $(a_0+1)^{2\beta} \geq (a_0+1)^{\beta}$, and the fourth step follows from simple algebra. 

{\bf Proof of Part 4.}

{\bf Proof of Part 4a.}

We can show
\begin{align*}
f_{\poly}(\A,x)_{j_0,j_1} 
= & ~ \langle  u_{\poly}(\A,x)_{j_0}, {\bf 1}_n \rangle^{-1} \cdot u_{\poly}(\A,x)_{j_0,j_1}\\ 
= & ~ \frac{ (a_0+1)^{\beta} }{ (a_0+1)^{\beta} + (n-1) \cdot 1 } \\
\leq & ~ \frac{1}{n^{1-c_0}}
\end{align*}
where the first step follows from the definition of $f_{\poly}$, the second step follows from {\bf Part 1} and {\bf Part 2}, and
and the last step follows from simple algebra.

{\bf Proof of Part 4b.}

We can show
\begin{align*}
f_{\poly}(\A,x)_{j_0,j_1} 
= & ~ \langle  u_{\poly}(\A,x)_{j_0}, {\bf 1}_n \rangle^{-1} \cdot u_{\poly}(\A,x)_{j_0,j_1}\\ 
= & ~ \frac{1}{ (a_0+1)^{\beta} + (n-1) \cdot 1 } \\
\leq & ~ \frac{1}{n}
\end{align*}
where the first step follows from the definition of $f_{\poly}$, the second step follows from {\bf Part 1} and {\bf Part 2}, and
and the last step follows from simple algebra.

\end{proof}

\subsubsection{The Property of Dataset 0 When Applying the Function \texorpdfstring{$c_{\poly}$}{}}
\label{sec:ds_0_c_exp}

In this section, we analyze the key properties of Dataset 0 while applying the function $c_{\poly}$.

\begin{lemma}\label{lem:self_dataset_0_c_exp}
If the following conditions hold
\begin{itemize}
    \item Let $\{A_1, A_2, A_3 \}$ from dataset ${\cal D}_0$ (see Definition~\ref{def:dataset_self_attention})
    \item Let $QK^\top = {\bf 1}_{d \times d}$
    \item Let $V = I_d$
    \item Let $n = t(d-2)$
\end{itemize}
Then for $c_{\poly}(\A,x)_{j_0,i_0} $ entry we have
\begin{itemize}
    \item {\bf Part 1.} For $j_0=j_3$ and $i_0 = 1$, we have $ c_{\poly}(\A,x)_{j_0,i_0}  \leq \frac{1}{n^{1-c_0}}a_0  $
    \item {\bf Part 2.} For $j_0=j_3$ and $i_0 \in \{2,\cdots,d-1\}$,
        we have $c_{\poly}(\A,x)_{j_0,i_0} \leq \frac{t}{n^{1-c_0}} b   $ 
    \begin{itemize}
        \item there is one index $i_0$ such that $ c_{\poly}(\A,x)_{j_0,i_0} \leq ( \frac{1}{ n^{1-c_0} } + \frac{t-1}{n}  ) b $ 
        \item the other indices $i_0$ are $c_{\poly}(\A,x)_{j_0,i_0} \leq \frac{t}{n} b  $
    \end{itemize}
    \item {\bf Part 3.} For $j_0 = j_3$ and $i_0 = d$, we have $ c_{\poly}(\A,x)_{j_0,i_0}  = c  $ 
    \item {\bf Part 4.} For $j_0 \neq j_3$ and $i_0 = 1$, we have $ c_{\poly}(\A,x)_{j_0,i_0} \leq \frac{1}{n^{1-c_0}} a_0  $
    \item {\bf Part 5.} For $j_0\neq j_3$ and $i_0 \in \{2,\cdots,d-1\}$, we have $ c_{\poly}(\A,x)_{j_0,i_0} \leq \frac{t}{n^{1-c_0}} b  $
    \begin{itemize}
        \item there is one index $i_0$ such that $c_{\poly}(\A,x)_{j_0,i_0} \leq ( \frac{1}{ n^{1-c_0} } + \frac{t-1}{n}  ) b$
        \item the other indices $i_0$ are $c_{\poly}(\A,x)_{j_0,i_0} \leq \frac{t}{n} b  $
    \end{itemize}
    \item {\bf Part 6.} For $j_0\neq j_3$ and $i_0 = d$, we have $ c_{\poly}(\A,x)_{j_0,i_0} = c $
\end{itemize}
\end{lemma}
\begin{proof}
{\bf Proof of Part 1.}

It follows from Part 3 of Lemma~\ref{lem:self_dataset_0_f_exp}, and Type I column in $A_3$.

{\bf Proof of Part 2.}

It follows from Part 3 of Lemma~\ref{lem:self_dataset_0_f_exp}, and Type II column in $A_3$.

There are two types of entries in $f_{\poly}(\A,x)_{j_0}$:
\begin{itemize}
    \item  there is one entry at most $1/n^{1-c_0}$
    \item there are $n-1$ entries at most $1/n$
\end{itemize} 

For $(A_3 V)_{i_0} \in \R^n$, the sparsity is $t$.

Thus there is one index $i_0$
\begin{align*}
\langle f_{\poly}(\A,x)_{j_0}, (A_3 V)_{i_0} \rangle \leq ( \frac{1}{ n^{1-c_0}} + \frac{t-1}{n} ) b.
\end{align*}
For the rest of indices $i_0$, we have
\begin{align*}
    \langle f_{\poly}(\A,x)_{j_0}, (A_3 V)_{i_0} \rangle \leq \frac{t}{n} b  .
\end{align*}

{\bf Proof of Part 3.}

It follows from Part 3 of Lemma~\ref{lem:self_dataset_0_f_exp}, and Type III column in $A_3$.

{\bf Proof of Part 4.}

It follows from Part 4 of Lemma~\ref{lem:self_dataset_0_f_exp}, and Type I column in $A_3$.

{\bf Proof of Part 5.}

It follows from Part 4 of Lemma~\ref{lem:self_dataset_0_f_exp}, and Type II column in $A_3$. The proof is similar to Part 2 of this Lemma.

{\bf Proof of Part 6.}

It follows from Part 4 of Lemma~\ref{lem:self_dataset_0_f_exp}, and Type III column in $A_3$.
\end{proof}

\subsubsection{The Property of Dataset 0 When Applying the Function \texorpdfstring{$c_{\poly}$}{} With Random Signs}
\label{sec:ds_0_y_exp}

In this section, we analyze the key properties of Dataset 0 while applying the function $c_{\poly}$ with random signs.

\begin{lemma}\label{lem:self_dataset_0_random_exp}
If the following conditions hold
\begin{itemize}
    \item Let $\{A_1, A_2,A_3,\}$ be from dataset ${\cal D}_0$ (see Definition~\ref{def:dataset_self_attention})
    \item Let $t \sqrt{d} = o(n^{1-c_0-0.01})$ (since $t(d-2) = n$, then this implies $d = \omega(n^{4 a_0+0.02})$)
\end{itemize}
Then, for each random string $\sigma \in \{-1,+1\}^d$, we have
\begin{itemize}
    \item {\bf Part 1.} If $j_0 = j_3$, then $\Pr[ |\langle c_{\poly}(\A,x)_{j_0} , \sigma \rangle| < (c+0.1)   ] \ge 1-\delta/\poly(n)$
    \item {\bf Part 2.} If $j_0 \neq j_3$, then $\Pr[ | \langle c_{\poly}(\A,x)_{j_0} , \sigma \rangle | < (c+0.1)   ] \ge 1 - \delta/\poly(n)$ 
\end{itemize}
\end{lemma}
\begin{proof}
{\bf Proof of Part 1.}
It follows from Part 1,2,3 of Lemma~\ref{lem:self_dataset_1_c_exp}, random sign distribution.

{\bf Proof of Part 2.}
It follows from Part 4,5,6 of Lemma~\ref{lem:self_dataset_1_c_exp} and Hoeffding inequality.

By Hoeffding inequality, we know that
\begin{align*}
| \langle c_{\poly}(\A,x)_{j_0} , \sigma \rangle - c  | \leq & ~ O( \sqrt{\log (n/\delta)} ) \cdot \frac{ t \sqrt{d} }{ n^{1-c_0} } b  \\
\leq & ~ 0.1  
\end{align*}
with probability at least $1-\delta/\poly(n)$. Here the last step due to $t\sqrt{d} = o(n^{1-c_0 - 0.01})$ and $\poly(\log n) \leq n^{0.01}$.
\end{proof}

\subsubsection{The Property of Dataset 0 When Applying the Function \texorpdfstring{$F_{\poly}$}{} }
\label{sec:ds_0_F_exp}

In this section, we analyze the key properties of Dataset 0 while applying the function $F_{\poly}$.

\begin{theorem}\label{thm:main:formal:part2}
If the following conditions hold
\begin{itemize}
    \item Let $d \in [ \omega(n^{4a_0 + 0.02}) , n ]$
    \item Let $\tau = (c+0.1)\sqrt{\log n}$
    \item Let $m = O(\log (n/\delta))$
    \item For any $\{A_1,A_2,A_3\}$ from ${\cal D}_1$ (Definition~\ref{def:dataset_self_attention})
    \item Let $F_{\poly}(A_1,A_2,A_3):=\phi( \sum_{j_0=1}^n \sum_{j_1=1}^m \phi_{\tau}( \langle c_{\poly}(\A,x)_{j_0} , y_{j_1} ) )$
\end{itemize}
Then we have
\begin{itemize}
    \item With high probability $1-\delta/\poly(n)$, $F_{\poly}(A_1,A_2,A_3) = 0$.  
\end{itemize}
\end{theorem}
\begin{proof}
It follows from using Lemma~\ref{lem:self_dataset_0_random_exp}.  
\end{proof}

\subsection{Dataset 1 When Applying \texorpdfstring{$F_{\poly}$}{} With Small \texorpdfstring{$\beta$}{}--A Low-Degree Polynomial Case}\label{sec:app:data_1_with_lin}
 
In Section~\ref{sec:ds_1_u_f_lin} we analyze the property of dataset 1 with respect to function $u_{\poly}$ and $f_{\poly}$. In Section~\ref{sec:ds_1_c_lin} we analyze the property of dataset 1 with respect to function $c_{\poly}$. In Section~\ref{sec:ds_1_y_lin} we analyze the property of dataset 1 with respect to function $c_{\poly}$ with random signs. In Section~\ref{sec:ds_1_F_lin} we show the property of dataset 1 with respect to the output of $F_{\poly}$.

\subsubsection{The Property of Dataset 1 When Applying Functions \texorpdfstring{$u_{\poly}$}{} and \texorpdfstring{$f_{\poly}$}{}}
\label{sec:ds_1_u_f_lin}

In this section, we analyze the key properties of Dataset 1 while applying the functions $u_{\poly}$ and $f_{\poly}$.

\begin{lemma}\label{lem:self_dataset_1_f_lin}
If the following conditions hold
\begin{itemize} 
    \item Let $b+c = 1$
    \item Let $c_0 \in (0,0.1)$
    \item Let $(1+a_1)^{\beta} < n^{c_0}$
    \item Let $\{ A_1, A_2, A_3 \}$ from dataset ${\cal D}_1$ (see Definition~\ref{def:dataset_self_attention})
    \item Let $QK^\top = {\bf 1}_{d \times d} $
\end{itemize} 
Then, for $u_{\poly}(\A,x)_{j_0,j_1}$ and $f_{\poly }(\A,x)_{j_0,j_1}$ entry we have

\begin{itemize}
    \item {\bf Part 1.} For $j_0 = j_3$, 
    \begin{itemize}
        \item {\bf Part 1a.} For $j_1 = j_3$, then  $u_{\poly}(\A,x)_{j_0,j_1} = (a_1 + 1 )^{2\beta} $.
        \item {\bf Part 1b.} For $j_1 \neq j_3$, then  $u_{\poly}(\A,x)_{j_0,j_1} = ( a_1 + 1 )^{\beta} $.
    \end{itemize}
    \item {\bf Part 2.} For $j_0 \neq j_3$
    \begin{itemize}
        \item {\bf Part 2a.} For $j_1 = j_3$, then $u_{\poly}(\A,x)_{j_0,j_1} = ( a_1 +1 )^{\beta}  $
        \item {\bf Part 2b.} For $j_1 \neq j_3$, then $u_{\poly}(A,x)_{j_0,j_1} = 1$.
    \end{itemize}
    \item {\bf Part 3.} For $j_0 = j_3$,
    \begin{itemize}
        \item {\bf Part 3a.} For $j_1 = j_3$, then  $f_{\poly}(\A,x)_{j_0,j_1} \leq \frac{1}{n^{1-c_0}}$
        \item {\bf Part 3b.} For $j_1 \neq j_3$, then  $f_{\poly}(\A,x)_{j_0,j_1} \leq \frac{1}{n}$
    \end{itemize}
    \item {\bf Part 4.} For $j_0 \neq j_3$,
    \begin{itemize}
        \item {\bf Part 4a.} For $j_1 = j_3$, then  $f_{\poly}(\A,x)_{j_0,j_1} \leq \frac{1}{n^{1-c_0}}$
        \item {\bf Part 4b.} For $j_1 \neq j_3$, then  $f_{\poly}(\A,x)_{j_0,j_1} \leq \frac{1}{n}$
    \end{itemize}
\end{itemize}

\end{lemma}
\begin{proof}

{\bf Proof of Part 1.}

{\bf Proof of Part 1a.}

For $j_0 = j_3$ and $j_1 = j_3$, by computing the circ product of two $3$-sparse vector
\begin{align*}
u_{\poly}(\A,x)_{j_0,j_1} 
= & ~ (a_1 +1)^{2\beta}
\end{align*}
where the first step follows from Definition \ref{def:dataset_self_attention}, the second step follows from simple algebra, and the last step follows from $c+b = 1$.

{\bf Proof of Part 1b.}

For $j_0 = j_3$ and $j_1 \neq j_3$.
\begin{align*}
u_{\poly}(\A,x)_{j_0,j_1} 
= & ~ (  a_1 + 1 )^{\beta}
\end{align*}
where the first step  follows from Definition \ref{def:dataset_self_attention}, the second step follows from simple algebra and the last step follows from $c+b = 1$.

{\bf Proof of Part 2.}

{\bf Proof of Part 2a.}

For $j_0 \neq j_3$ and $j_1 = j_3$, this case is same as Part 1b.

{\bf Proof of Part 2b.}
For $j_0 \neq j_3$ and $j_1 \neq j_3$,

For the case that we tensor a $2$-sparse vector with another $2$-sparse vector, we have 
\begin{align*}
u_{\poly}(\A,x)_{j_0,j_1} 
= & ~ 1
\end{align*}
where the first step follows from Definition \ref{def:dataset_self_attention}, the second step follows from simple algebra and the last step follows from $b+c = 1$.

{\bf Proof of Part 3.}

{\bf Proof of Part 3a.}

We can show that
\begin{align*}
\langle u_{\poly}(\A,x)_{j_0} , {\bf 1}_n \rangle = (a_1+1)^{2\beta} + (n-1) \cdot (a_1 + 1)^{\beta}.
\end{align*}
The reason is $u_{\poly}(\A,x)_{j_0} $ is a length-$n$ vector where $n-1$ coordinates are $(a_1+1)^{\beta}$ and only coordinate is $(a_1+1)^{2\beta}$.

Using Part 1a, we have
\begin{align*}
u_{\poly}(\A,x)_{j_0,j_1} = (a_1 + 1)^{2\beta}
\end{align*}

We can show
\begin{align*}
f_{\poly}(\A,x)_{j_0,j_1} = & ~ \langle  u_{\poly}(\A,x)_{j_0}, {\bf 1}_n \rangle^{-1} \cdot u_{\poly}(\A,x)_{j_0,j_1}\\ 
= & ~ \frac{1}{ (a_1+1)^{2\beta} + (n-1) (a_1 + 1)^{\beta} } \cdot (a_1+1)^{2\beta} \\
\leq & ~ \frac{1}{ (a_1+1)^{\beta} + (n-1) (1+a_1)^{\beta} } \cdot  (a_1+1)^{2\beta} \\
= & ~ \frac{(a_1+1)^{\beta} }{ n  } \\
\leq & ~ \frac{1}{n^{1-c_0}}
\end{align*}
where the first step follows from the definition of $f_{\poly}$, the second step follows from  {\bf Part 1}, the third step follows from simple algebra and the last step follows from simple algebra.

{\bf Proof of Part 3b.}
\begin{align*}
f_{\poly}(\A,x)_{j_0,j_1} = & ~ \langle  u_{\poly}(\A,x)_{j_0}, {\bf 1}_n \rangle^{-1} \cdot u_{\poly}(\A,x)_{j_0,j_1}\\ 
= & ~ \frac{1}{ (a_1+1)^{2\beta}   + (n-1) (a_1+1)^{\beta} } \cdot  (a_1+1)^{\beta} \\
\leq & ~ 1/n
\end{align*}
where the first step follows from the definition of $f_{\poly}$, the second step follows from  {\bf Part 1}, the third step follows from simple algebra and the last step follows from simple algebra. 
{\bf Proof of Part 4.}

{\bf Proof of Part 4a.}
We can show
\begin{align*}
f_{\poly}(\A,x)_{j_0,j_1} \leq & ~ \frac{1}{n^{1-c_0}}
\end{align*}

{\bf Proof of Part 4b.}

We can show
\begin{align*}
f_{\poly}(\A,x)_{j_0,j_1} \leq & ~\frac{1}{n}
\end{align*}

\end{proof}

\subsubsection{The Property of Dataset 1 When Applying the Function \texorpdfstring{$c_{\poly}$}{}}
\label{sec:ds_1_c_lin}

In this section, we analyze the key properties of Dataset 1 while applying the function $c_{\poly}$.

\begin{lemma}\label{lem:self_dataset_1_c_lin}
If the following conditions hold
\begin{itemize}
    \item Let $\{A_1, A_2, A_3 \}$ from dataset ${\cal D}_1$ (see Definition~\ref{def:dataset_self_attention})
    \item Let $QK^\top = {\bf 1}_{d \times d}$
    \item Let $V = I_d$
    \item Let $n = t(d-2)$
\end{itemize}
Then for $c_{\poly}(\A,x)_{j_0,i_0} $ entry we have
\begin{itemize}
    \item {\bf Part 1.} For $j_0=j_3$ and $i_0 = 1$, we have $ c_{\poly}(\A,x)_{j_0,i_0}  \leq \frac{ 1 }{n^{1-c_0} }  \cdot a_1 $
    \item {\bf Part 2.} For $j_0=j_3$ and $i_0 \in \{2,\cdots,d-1\}$,
        we have $c_{\poly}(\A,x)_{j_0,i_0} \leq \frac{1}{n^{1-c_0}} \cdot t \cdot b  $ 
    \item {\bf Part 3.} For $j_0 = j_3$ and $i_0 = d$, we have $ c_{\poly}(\A,x)_{j_0,i_0} = c $ 
    \item {\bf Part 4.} For $j_0 \neq j_3$ and $i_0 = 1$, we have $ c_{\poly}(\A,x)_{j_0,i_0} \leq \frac{1}{n^{1-c_0}} \cdot a_1 $
    \item {\bf Part 5.} For $j_0\neq j_3$ and $i_0 \in \{2,\cdots,d-1\}$, we have $ c_{\poly}(\A,x)_{j_0,i_0} \leq \frac{1}{n^{1-c_0}} \cdot t \cdot b $
    \item {\bf Part 6.} For $j_0\neq j_3$ and $i_0 = d$, we have $ c_{\poly}(\A,x)_{j_0,i_0} = c  $
\end{itemize}
\end{lemma}
\begin{proof}
{\bf Proof of Part 1.}

It follows from Part 3 of Lemma~\ref{lem:self_dataset_1_f_lin}, and Type I column in $A_3$.

{\bf Proof of Part 2.}

It follows from Part 3 of Lemma~\ref{lem:self_dataset_1_f_lin}, and Type II column in $A_3$.

We know that each entry in $f_{\poly}(\A,x)_{j_0}$ is at least $0$ and is at most $\frac{1}{n^{1-c_0}}$.

We know that each entry in $(A_3 V)_{i_0} \in \R^n$ is at least $0$ and at most $b $ and it is $t$-sparse.

Thus,
\begin{align*}
\langle f_{\poly}(\A,x)_{j_0}, (A_3 V)_{i_0} \rangle \leq \frac{t}{n^{1-c_0}} b 
\end{align*}

{\bf Proof of Part 3.}

It follows from Part 3 of Lemma~\ref{lem:self_dataset_1_f_lin}, and Type III column in $A_3$.

{\bf Proof of Part 4.}

It follows from Part 4 of Lemma~\ref{lem:self_dataset_1_f_lin}, and Type I column in $A_3$.

{\bf Proof of Part 5.}

It follows from Part 4 of Lemma~\ref{lem:self_dataset_1_f_lin}, and Type II column in $A_3$.

{\bf Proof of Part 6.}

It follows from Part 4 of Lemma~\ref{lem:self_dataset_1_f_lin}, and Type III column in $A_3$.
\end{proof}

\subsubsection{The Property of Dataset 1 When Applying the Function \texorpdfstring{$c_{\poly}$}{} With Random Signs}
\label{sec:ds_1_y_lin}

In this section, we analyze the key properties of Dataset 1 while applying the function $c_{\poly}$ with random signs.

\begin{lemma}\label{lem:self_dataset_1_random_lin}
If the following conditions hold
\begin{itemize}
    \item Let $\{A_1, A_2,A_3,\}$ be from dataset ${\cal D}_1$ (see Definition~\ref{def:dataset_self_attention})
    \item Let $t \sqrt{d} = o(n^{0.99})$ (since $t(d-2) = n$, then this implies $d = \omega(n^{0.02})$)
\end{itemize}
Then, for each random string $\sigma \in \{-1,+1\}^d$, we have
\begin{itemize}
    \item {\bf Part 1.} If $j_0 = j_3$, then $\Pr[ |\langle c_{\poly}(\A,x)_{j_0} , \sigma \rangle| < (c+0.1) \sqrt{\log n} ] \geq 1-\delta/\poly(n)$
    \item {\bf Part 2.} If $j_0 \neq j_3$, then $\Pr[ | \langle c_{\poly}(\A,x)_{j_0} , \sigma \rangle | < (c+0.1) \sqrt{\log n} ] \geq 1 - \delta/\poly(n)$
\end{itemize}
\end{lemma}
\begin{proof}
{\bf Proof of Part 1.}
It follows from Part 1,2,3 of Lemma~\ref{lem:self_dataset_1_c_lin} and Hoeffding inequality (Lemma~\ref{lem:hoeffding_bound}).

By hoeffding inequality (Lemma~\ref{lem:hoeffding_bound}), we know that
\begin{align*}
| \langle c_{\poly}(\A,x)_{j_0} , \sigma \rangle - c \sqrt{\log n} | \leq & ~ O( \sqrt{\log (n/\delta)} ) \cdot \frac{  t \sqrt{d} }{ n^{1-c_0} } b  \\
\leq & ~ 0.1  
\end{align*}
with probability at least $1-\delta/\poly(n)$. Here the last step due to $t\sqrt{d} = o(n^{0.99})$, $a_1 = O(1)$, $b = O(1)$, and $\poly(\log n) \leq n^{0.01}$.

{\bf Proof of Part 2.}
It follows from Part 4,5,6 of Lemma~\ref{lem:self_dataset_1_c_lin} and Hoeffding inequality (Lemma~\ref{lem:hoeffding_bound}).

By hoeffding inequality (Lemma~\ref{lem:hoeffding_bound}), we know that
\begin{align*}
| \langle c_{\poly}(\A,x)_{j_0} , \sigma \rangle - c \sqrt{\log n} | \leq & ~ O( \sqrt{\log (n/\delta)} ) \cdot \frac{  t \sqrt{d} }{ n^{1-c_0} } b  \\
\leq & ~ 0.1  
\end{align*}
with probability at least $1-\delta/\poly(n)$. Here the last step due to $t\sqrt{d} = o(n^{0.99})$, $a_1 = O(1)$, $b = O(1)$, and $\poly(\log n) \leq n^{0.01}$.
\end{proof}

\subsubsection{The Property of Dataset 1 When Applying the Function \texorpdfstring{$F_{\poly}$}{}}
\label{sec:ds_1_F_lin}

In this section, we analyze the key properties of Dataset 1 while applying the function $F_{\poly}$.

\begin{theorem}\label{thm:main:formal:part3}
If the following conditions hold
\begin{itemize}
    \item Let $d \in [ \omega(n^{0.02}) , n ]$
    \item Let $\tau = (c+0.1)$
    \item Let $m = O(\log (n/\delta))$
    \item For any $\{A_1,A_2,A_3\}$ from ${\cal D}_1$ (Definition~\ref{def:dataset_self_attention})
    \item Let $F_{\poly}(A_1,A_2,A_3):=\phi( \sum_{j_0=1}^n \sum_{j_1=1}^m \phi_{\tau}( \langle c_{\poly}(\A,x)_{j_0} , y_{j_1} ) )$
\end{itemize}
Then we have
\begin{itemize}
    \item With high probability $1-\delta/\poly(n)$, $F_{\poly}(A_1,A_2,A_3) = 0$
\end{itemize}
\end{theorem}
\begin{proof}
It follows from using Lemma~\ref{lem:self_dataset_1_random_exp}.
\end{proof}

\subsection{Dataset 0 When Applying \texorpdfstring{$F_{\poly}$}{} With Small \texorpdfstring{$\beta$}{}--A Low-Degree Polynomial Case}
\label{sec:app:data_0_with_lin}
 
In Section~\ref{sec:ds_0_u_f_lin} we analyse the property of dataset 0 with respect to function $u_{\poly}$ and $f_{\poly}$. In Section~\ref{sec:ds_0_c_lin} we analyse the property of dataset 0 with respect to function $c_{\poly}$. In Section~\ref{sec:ds_0_y_lin} we analyse the property of dataset 0 with respect to function $c_{\poly}$ with random signs. In Section~\ref{sec:ds_0_F_lin} we show the property of dataset 0 with respect to the output of $F_{\poly}$.

\subsubsection{The Property of Dataset 0 When Applying Functions \texorpdfstring{$u_{\poly}$}{} and \texorpdfstring{$f_{\poly}$}{}}
\label{sec:ds_0_u_f_lin}

In this section, we analyze the key properties of Dataset 0 while applying the function $u_{\poly}$ and $f_{\poly}$.

\begin{lemma}\label{lem:self_dataset_0_f_lin}
If the following conditions hold
\begin{itemize}
    \item Let $\{ A_1, A_2, A_3 \}$ from dataset ${\cal D}_0$ (see Definition~\ref{def:dataset_self_attention})
    \item Let $QK^\top = I_d $
    \item Let $c_0 \in (0,0.1)$
    \item Let $(a_0+1)^{\beta} < n^{c_0}$
\end{itemize} 
Then, for $u_{\poly}(\A,x)_{j_0,j_1}$ and $f_{\poly}(\A,x)_{j_0,j_1}$ entry we have

\begin{itemize}
    \item {\bf Part 1.} For $j_0 = j_3$, 
    \begin{itemize}
        \item {\bf Part 1a.} For $j_1 = j_3$, then  $u_{\poly}(\A,x)_{j_0,j_1} = (a_0 + 1)^{2\beta}$.
        \item {\bf Part 1b.} For $j_1 \neq j_3$, then  $u_{\poly}(\A,x)_{j_0,j_1} = (a_0 + 1)^{\beta}  $.
    \end{itemize}
    \item {\bf Part 2.} For $j_0 \neq j_3$
    \begin{itemize}
        \item {\bf Part 2a.} For $j_1 = j_3$, then $u_{\poly}(\A,x)_{j_0,j_1} = (a_0+1)^{\beta}  $
        \item {\bf Part 2b.} For $j_1 \neq j_3$, then $u_{\poly}(A,x)_{j_0,j_1} = 1$.
    \end{itemize}
    \item {\bf Part 3.} For $j_0 = j_3$,
    \begin{itemize}
        \item {\bf Part 3a.} For $j_1 = j_3$, then  $f_{\poly}(\A,x)_{j_0,j_1} \leq \frac{1}{n^{1-c_0}}$
        \item {\bf Part 3b.} For $j_1 \neq j_3$, then  $f_{\poly}(\A,x)_{j_0,j_1} \leq \frac{1}{n}$
    \end{itemize}
    \item {\bf Part 4.} For $j_0 \neq j_3$,
    \begin{itemize}
        \item {\bf Part 4a.} For $j_1 = j_3$, then  $f_{\poly}(\A,x)_{j_0,j_1} \leq \frac{1}{n^{1-c_0}}$
        \item {\bf Part 4b.} For $j_0 \neq j_3$, then  $f_{\poly}(\A,x)_{j_0,j_1} \leq \frac{1}{n}$
    \end{itemize}
\end{itemize}

\end{lemma}

\begin{proof}

{\bf Proof of Part 1.}

{\bf Proof of Part 1a.}

For $j_0 = j_3$ and $j_1 = j_3$, by computing the circ product of two $3$-sparse vector
\begin{align*}
u_{\poly}(\A,x)_{j_0,j_1} 
= & ~ ( a_0 + 1 )^{2\beta}
\end{align*}
where the first step  follows from Definition \ref{def:dataset_self_attention}, the second step follows from simple algebra and the last step follows from $c+b = 1$.

{\bf Proof of Part 1b.}

For $j_0 = j_3$ and $j_1 \neq j_3$.
\begin{align*}
u_{\poly}(\A,x)_{j_0,j_1} 
= & ~ ( a_0 + 1 )^{\beta}
\end{align*}
where the first step  follows from Definition \ref{def:dataset_self_attention}, the second step follows from simple algebra and the last step follows from $c+b = 1$.

{\bf Proof of Part 2.}

{\bf Proof of Part 2a.}

For $j_0 \neq j_3$ and $j_1 = j_3$, this case is same as Part 1b.

{\bf Proof of Part 2b.}
For $j_0 \neq j_3$ and $j_1 \neq j_3$,

For the case that we tensor a $2$-sparse vector with another $2$-sparse vector, we have 
\begin{align*}
u_{\poly}(\A,x)_{j_0,j_1} 
= & ~ 1
\end{align*}
where the first step follows from Definition \ref{def:dataset_self_attention}, the second step follows from simple algebra and the last step follows from $b+c = 1$.

{\bf Proof of Part 3.}

{\bf Proof of Part 3a.}

We can show
\begin{align*}
f_{\poly}(\A,x)_{j_0,j_1} = & ~ \langle  u_{\poly}(\A,x)_{j_0}, {\bf 1}_n \rangle^{-1} \cdot u_{\poly}(\A,x)_{j_0,j_1}\\ 
\leq & ~ \frac{1}{ n^{1-c_0} }
\end{align*}
where the first step follows from the definition of $f_{\poly}$, the second step follows from  {\bf Part 1}, the third step follows from simple algebra and the last step follows from simple algebra. 

{\bf Proof of Part 3b.}
\begin{align*}
f_{\poly}(\A,x)_{j_0,j_1} = & ~ \langle  u_{\poly}(\A,x)_{j_0}, {\bf 1}_n \rangle^{-1} \cdot u_{\poly}(\A,x)_{j_0,j_1}\\ 
\leq & ~ 1/n
\end{align*}
where the first step follows from the definition of $f_{\poly}$, the second step follows from  {\bf Part 1}, the third step follows from simple algebra and the last step follows from simple algebra. 
{\bf Proof of Part 4.}

{\bf Proof of Part 4a.}
We can show
\begin{align*}
f_{\poly}(\A,x)_{j_0,j_1} \leq & ~ \frac{1}{n^{1-c_0}}
\end{align*}

{\bf Proof of Part 4b.}

We can show
\begin{align*}
f_{\poly}(\A,x)_{j_0,j_1} \leq & ~  \frac{1}{n}
\end{align*}

\end{proof}

\subsubsection{The Property of Dataset 0 When Applying the Function \texorpdfstring{$c_{\poly}$}{}}
\label{sec:ds_0_c_lin}

In this section, we analyze the key properties of Dataset 0 while applying the function $c_{\poly}$.

\begin{lemma}\label{lem:self_dataset_0_c_lin}
If the following conditions hold
\begin{itemize}
    \item Let $\{A_1, A_2, A_3 \}$ from dataset ${\cal D}_0$ (see Definition~\ref{def:dataset_self_attention})
    \item Let $QK^\top = {\bf 1}_{d \times d}$
    \item Let $V = I_d$
    \item Let $n = t(d-2)$
\end{itemize}
Then for $c_{\poly}(\A,x)_{j_0,i_0} $ entry we have
\begin{itemize}
    \item {\bf Part 1.} For $j_0=j_3$ and $i_0 = 1$, we have $ c_{\poly}(\A,x)_{j_0,i_0}  \leq \frac{ (1+a_0)}{n}  \cdot a_0 $
    \item {\bf Part 2.} For $j_0=j_3$ and $i_0 \in \{2,\cdots,d-1\}$,
        we have $c_{\poly}(\A,x)_{j_0,i_0} \leq \frac{(1+a_0)}{n} \cdot t \cdot b    $ 
    \item {\bf Part 3.} For $j_0 = j_3$ and $i_0 = d$, we have $ c_{\poly}(\A,x)_{j_0,i_0} = c  $ 
    \item {\bf Part 4.} For $j_0 \neq j_3$ and $i_0 = 1$, we have $ c_{\poly}(\A,x)_{j_0,i_0} \leq \frac{(1+a_0)}{n} \cdot a_0  $
    \item {\bf Part 5.} For $j_0\neq j_3$ and $i_0 \in \{2,\cdots,d-1\}$, we have $ c_{\poly}(\A,x)_{j_0,i_0} \leq \frac{(1+a_0)}{n} \cdot t \cdot b  $
    \item {\bf Part 6.} For $j_0\neq j_3$ and $i_0 = d$, we have $ c_{\poly}(\A,x)_{j_0,i_0} = c $
\end{itemize}
\end{lemma}
\begin{proof}
{\bf Proof of Part 1.}

It follows from Part 3 of Lemma~\ref{lem:self_dataset_0_f_lin}, and Type I column in $A_3$.

{\bf Proof of Part 2.}

It follows from Part 3 of Lemma~\ref{lem:self_dataset_0_f_lin}, and Type II column in $A_3$.

We know that each entry in $f_{\poly}(\A,x)_{j_0}$ is at least $0$ and is at most $\frac{1+a_0}{n}$.

We know that each entry in $(A_3 V)_{i_0} \in \R^n$ is at least $0$ and at most $b \sqrt{ \log n}$ and it is $t$-sparse.

Thus,
\begin{align*}
\langle f_{\poly}(\A,x)_{j_0}, (A_3 V)_{i_0} \rangle \leq \frac{t}{n^{1-c_0} } b  
\end{align*}

{\bf Proof of Part 3.}

It follows from Part 3 of Lemma~\ref{lem:self_dataset_0_f_lin}, and Type III column in $A_3$.

{\bf Proof of Part 4.}

It follows from Part 4 of Lemma~\ref{lem:self_dataset_0_f_lin}, and Type I column in $A_3$.

{\bf Proof of Part 5.}

It follows from Part 4 of Lemma~\ref{lem:self_dataset_0_f_lin}, and Type II column in $A_3$.

{\bf Proof of Part 6.}

It follows from Part 4 of Lemma~\ref{lem:self_dataset_0_f_lin}, and Type III column in $A_3$.
\end{proof}

\subsubsection{The Property of Dataset 0 When Applying the Function \texorpdfstring{$c_{\poly}$}{} With Random Signs}
\label{sec:ds_0_y_lin}

In this section, we analyze the key properties of Dataset 0 while applying the function $c_{\poly}$ with random signs.

\begin{lemma}\label{lem:self_dataset_0_random_lin}
If the following conditions hold
\begin{itemize}
    \item Let $\{A_1, A_2,A_3,\}$ be from dataset ${\cal D}_0$ (see Definition~\ref{def:dataset_self_attention})
    \item Let $t \sqrt{d} = o(n^{0.99})$ (since $t(d-2) = n$, then this implies $d = \omega(n^{0.02})$)
\end{itemize}
Then, for each random string $\sigma \in \{-1,+1\}^d$, we have
\begin{itemize}
    \item {\bf Part 1.} If $j_0 = j_3$, then $\Pr[ |\langle c_{\poly}(\A,x)_{j_0} , \sigma \rangle| < (c+0.1) \sqrt{\log n} ] \geq 1-\delta/\poly(n)$
    \item {\bf Part 2.} If $j_0 \neq j_3$, then $\Pr[ | \langle c_{\poly}(\A,x)_{j_0} , \sigma \rangle | < (c+0.1) \sqrt{\log n} ] \geq 1 - \delta/\poly(n)$
\end{itemize}
\end{lemma}
\begin{proof}
{\bf Proof of Part 1.}
It follows from Part 1,2,3 of Lemma~\ref{lem:self_dataset_0_c_lin} and Hoeffding inequality (Lemma~\ref{lem:hoeffding_bound}).

By hoeffding inequality (Lemma~\ref{lem:hoeffding_bound}), we know that
\begin{align*}
| \langle c_{\poly}(\A,x)_{j_0} , \sigma \rangle - c   | \leq & ~ O( \sqrt{\log (n/\delta)} ) \cdot \frac{  t \sqrt{d} }{ n^{1-c_0} } b   \\
\leq & ~ 0.1  
\end{align*}
with probability at least $1-\delta/\poly(n)$. Here the last step due to $t\sqrt{d} = o(n^{0.99})$, $a_0 = O(1)$, $b = O(1)$, and $\poly(\log n) \leq n^{0.01}$.

{\bf Proof of Part 2.}
It follows from Part 4,5,6 of Lemma~\ref{lem:self_dataset_0_c_lin} and Hoeffding inequality (Lemma~\ref{lem:hoeffding_bound}).

By hoeffding inequality (Lemma~\ref{lem:hoeffding_bound}), we know that
\begin{align*}
| \langle c_{\poly}(\A,x)_{j_0} , \sigma \rangle - c   | \leq & ~ O( \sqrt{\log (n/\delta)} ) \cdot \frac{   t \sqrt{d} }{ n^{1-c_0} } b \sqrt{ \log n} \\
\leq & ~ 0.1 
\end{align*}
with probability at least $1-\delta/\poly(n)$. Here the last step due to $t\sqrt{d} = o(n^{0.99})$, $a_0 = O(1)$, $b = O(1)$, and $\poly(\log n) \leq n^{0.01}$.
\end{proof}
\subsubsection{The Property of Dataset 0 When Applying the Function \texorpdfstring{$F_{\poly}$}{} }
\label{sec:ds_0_F_lin}

In this section, we analyze the key properties of Dataset 0 while applying the function $F_{\poly}$.

\begin{theorem}\label{thm:main:formal:part4}
If the following conditions hold
\begin{itemize}
    \item Let $d \in [ \omega(n^{0.02}) , n ]$
    \item Let $\tau = (c+0.1) $
    \item Let $m = O(\log (n/\delta))$
    \item For any $\{A_1,A_2,A_3\}$ from ${\cal D}_0$ (Definition~\ref{def:dataset_self_attention})
    \item Let $F_{\poly}(A_1,A_2,A_3):=\phi( \sum_{j_0=1}^n \sum_{j_1=1}^m \phi_{\tau}( \langle c_{\poly}(\A,x)_{j_0} , y_{j_1} ) )$
\end{itemize}
Then we have
\begin{itemize}
    \item With high probability $1-\delta/\poly(n)$, $F_{\poly}(A_1,A_2,A_3) = 0$
\end{itemize}
\end{theorem}
\begin{proof}
It follows from using Lemma~\ref{lem:self_dataset_0_random_lin}.
\end{proof}

\ifdefined\isarxiv
\bibliographystyle{alpha}
\bibliography{ref}
\else
\bibliography{ref}
\bibliographystyle{alpha}

\fi

\newpage
\onecolumn
\appendix




\end{document}